\documentclass{article}

\usepackage{microtype}
\usepackage{graphicx}
\usepackage{booktabs} 

\usepackage{microtype}
\usepackage{graphicx}
\usepackage{subfigure}
\usepackage{caption}
\usepackage{booktabs} 
\usepackage[pagebackref=false,breaklinks=true,colorlinks,bookmarks=false]{hyperref}
\hypersetup{linkcolor=[rgb]{0.7,0.1,0.1}}
\hypersetup{citecolor=[rgb]{0.4,0.15,0.95}}

\usepackage{algorithm}
\usepackage{algorithmic}
\renewcommand{\algorithmiccomment}[1]{/* #1 */}

\usepackage{colortbl}

\usepackage{amsmath}
\usepackage{amssymb}
\usepackage{mathtools}
\usepackage{amsthm}

\usepackage[capitalize,noabbrev]{cleveref}
\usepackage{colortbl}
\usepackage{nicematrix}
\usepackage[inline, shortlabels]{enumitem}
\usepackage{multirow}

\usepackage{amsthm}
\usepackage{amsmath,amsfonts,bm}
\usepackage{nicefrac}
\usepackage{array}
\usepackage{wrapfig}

\usepackage{url}
\usepackage{pbox}

\definecolor{bluex}{rgb}{0.27, 0.42, 0.81}
\definecolor{purplex}{HTML}{9564bf}
\definecolor{red3}{HTML}{C52A20}
\definecolor{red2}{HTML}{B36A6F}
\definecolor{red1}{HTML}{FFb5b5}
\definecolor{purple}{HTML}{B36A6F}
\definecolor{darkyellow}{HTML}{D5BA82}
\definecolor{blue1}{HTML}{508AB2}
\definecolor{blue2}{HTML}{C4E4E3}
\definecolor{green1}{HTML}{A1D0C7}
\definecolor{green2}{HTML}{BFF6BA}
\definecolor{green3}{HTML}{028100}
\definecolor{teal}{HTML}{508AB2}
\definecolor{Gray}{gray}{0.94}
\definecolor{Gray2}{gray}{0.7}
\definecolor{orange3}{HTML}{c28c69}
\definecolor{blue3}{HTML}{3b75af}

\usepackage{pifont}

\usepackage[listings]{tcolorbox}
\tcbuselibrary{listings,theorems}
\newtcolorbox{mybox}{colback=white!5!white,colframe=black!75!black, left=.05in, right=.05in}

\newtcbtheorem{emptybox}{}%
{colback=green2!5,colframe=blue1,fonttitle=\bfseries,theorem style=plain, left=.0in, right=.0in,bottom=.02in, top=.02in,terminator sign none}{emptybox}
\newtcbtheorem[]{emptyboxx}{A Simple Template For Creating Backward Question}%
{colback=green2!5,colframe=blue1,fonttitle=\bfseries,theorem style=plain, left=.0in, right=.0in,bottom=.02in, top=.02in,terminator sign none}{emptyboxx}
\newtcbtheorem{bth}{Theoreme}
{colback=red!20,colframe=red,theorem style=plain,fonttitle=\bfseries,coltitle=black,terminator sign none}{th}

\newtcbtheorem[]{exmp}{Example}%
{colback=green2!5,colframe=blue1,fonttitle=\bfseries, left=.02in, right=.02in,bottom=.02in, top=.02in}{exmp}
\newtcbtheorem[]{template}{Template}%
{colback=green2!5,colframe=blue1,fonttitle=\bfseries, left=.02in, right=.02in,bottom=.02in, top=.02in}{template}
\newtcbtheorem[]{prompt}{Prompt}%
{colback=green2!5,colframe=blue1,fonttitle=\bfseries, left=.02in, right=.02in,bottom=.02in, top=.02in}{prompt}
\newtcbtheorem[number within=section]{thm}{Theorem}%
{colback=green!5,colframe=green!35!black,fonttitle=\bfseries}{th}
\newtcbtheorem[number within=section]{corr}{Corollary}%
{colback=green!5,colframe=green!35!black,fonttitle=\bfseries}{th}
\newtcbtheorem{method}{Method}%
{colback=green!5,colframe=green!35!black,fonttitle=\bfseries}{th}

\newtcbtheorem{ebox}{}
{colback=green2!5,colframe=blue1, left=.02in, right=.02in,bottom=.02in,title empty, top=.02in, theorem style=plain apart,frame empty}{ebox}
\theoremstyle{plain}

\theoremstyle{definition}

\theoremstyle{remark}

\DeclareMathOperator*{\argmax}{arg\,max}

\newcommand{\bP}{\mathbb{P}}

\newcommand{\hX}{\mathcal{X}}

\newcommand{\hB}{\mathcal{B}}
\newcommand{\hC}{\mathcal{C}}
\newcommand{\hD}{\mathcal{D}}

\newcommand{\hK}{\mathcal{K}}
\newcommand{\hL}{\mathcal{L}}

\newcommand{\vg}{{\bf g}}

\newcommand{\vv}{{\bf v}}
\newcommand{\vw}{{\bf w}}
\newcommand{\vx}{{\bf x}}

\newcommand{\vC}{{\bf C}}

\newcommand{\vI}{{\bf I}}

\newcommand{\vL}{{\bf L}}

\newcommand{\vV}{{\bf V}}
\newcommand{\vW}{{\bf W}}

\DeclareMathOperator{\ssim}{\textsf{sim}}

\newcommand{\vtheta}{{\boldsymbol \theta}}

\usepackage{hyperref}
\usepackage{url}

\usepackage[accepted]{icml2026}

\usepackage{amsmath}
\usepackage{amssymb}
\usepackage{mathtools}
\usepackage{amsthm}

\usepackage[capitalize,noabbrev]{cleveref}

\usepackage[textsize=tiny]{todonotes}

\icmltitlerunning{SPARD: Defending Harmful Fine-Tuning Attack via Safety Projection with
Relevance–Diversity Data Selection}

\begin{document}

\twocolumn[
  \icmltitle{SPARD: Defending Harmful Fine-Tuning Attack via \\ Safety Projection with
Relevance–Diversity Data Selection}




  \icmlsetsymbol{equal}{*}

  \begin{icmlauthorlist}
    \icmlauthor{Shuhao Chen}{sustech,hkust}
    \icmlauthor{Weisen Jiang}{cuhk}
    \icmlauthor{Yeqi Gong}{pcg}
    \icmlauthor{Shengda Luo}{cmgl}
    \icmlauthor{Chengxiang Zhuo}{pcg}
    \icmlauthor{Zang Li}{pcg}
    \icmlauthor{James T. Kwok}{hkust}
    \icmlauthor{Yu Zhang}{sustech}
  \end{icmlauthorlist}

  \icmlaffiliation{hkust}{Department of Computer Science and Engineering, The Hong Kong University of Science and Technology}
  \icmlaffiliation{sustech}{Department of Computer Science and Engineering, Southern University of Science and Technology}
  \icmlaffiliation{cuhk}{Department of Computer Science and Engineering, The Chinese University of Hong Kong}
  \icmlaffiliation{pcg}{Platform and Content Group, Tencent}
  \icmlaffiliation{cmgl}{Chinese Medicine Guangdong Laboratory}

  \icmlcorrespondingauthor{Yu Zhang}{yu.zhang.ust@gmail.com}

  \icmlkeywords{Machine Learning, ICML}

  \vskip 0.3in
]



\printAffiliationsAndNotice{}  

\begin{abstract}
Fine-tuning large language models often undermines their safety alignment, a problem further amplified by harmful fine-tuning attacks in which adversarial data removes safeguards and induces unsafe behaviors. 
We propose \textbf{SPARD}, a novel defense framework that integrates \textbf{S}afety-\textbf{P}rojected \textbf{A}lternating optimization with \textbf{R}elevance-\textbf{D}iversity aware data selection. 
SPARD optimizes alternatively between utility updates and explicit safety projections with a set of safe data to enforce safety constraints.
To curate safe data, we propose a Relevance–Diversity Determinantal Point Process to select compact safe data, balancing task relevance and safety coverage. 
Experiments on GSM8K and OpenBookQA under four harmful fine-tuning attacks demonstrate that SPARD consistently achieves the lowest average attack success rates, substantially outperforming state-of-the-art defense methods, while maintaining high task accuracy. Code is available at \url{https://github.com/shuhao02/SPARD}.
\end{abstract}

\section{Introduction}

Large language models (LLMs)~\citep{gpt4, touvron2023llama2, yang2024qwen, llama3-2, jiang2024forward} 
have shown strong capabilities across a wide range of tasks, making them increasingly popular in real-world applications~\citep{jiang2023effective, wei2024gita, chen2024routerdc}.  
Fine-tuning-as-a-service has become a common way for users to adapt LLMs to specific downstream domains via service providers.
However, fine-tuning can inadvertently undermine safety alignment, causing models to forget their safeguards~\citep{qi2024finetuning, yang2023shadow, lermen2023lora}.  
This problem becomes more severe when fine-tuning data contains malicious or adversarial content, as in harmful fine-tuning attacks~\citep{liu2023jailbreaking, zou2023universal, huang2024harmful, jiang2026metadefense}, which can effectively strip away safety mechanisms and cause the model to produce unsafe outputs.  

\begin{figure}[t]
    \centering
    \includegraphics[width=0.95\linewidth]{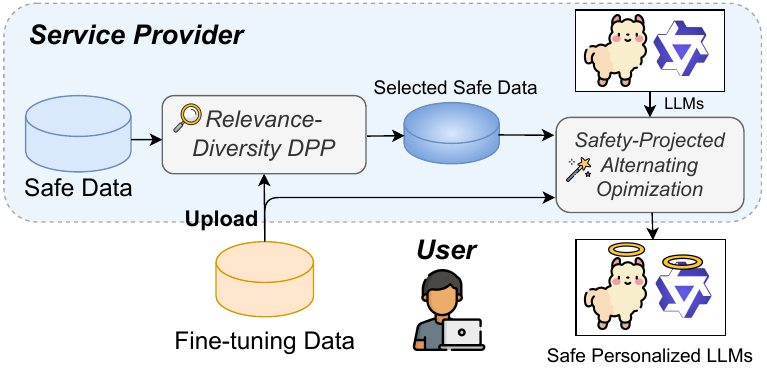}
    \vspace{-.05in}
    \caption{Illustration of SPARD.}
    \vspace{-.25in}
    \label{fig:SPARD_illustration}
\end{figure}

Recently, many defense methods have been proposed to counter the harmful fine-tuning attacks. 
For example, 
PTST~\citep{lyu2024keeping} and SafeLoRA~\citep{hsu2024safe} mitigate harmful updates by re-injecting safety prompts or constraining LoRA adapters, but they either rely on carefully crafted prompt templates or require structural constraints that limit general applicability. 
Other approaches exploit safe data as an implicit safeguard. 
For instance, SafeInstr~\citep{bianchi2023safety} mixes a small fraction of safe examples into fine-tuning data to counter harmful behaviors, 
while Lisa~\citep{huang2024lisa} uses a bi-state optimization with safe samples and applies a proximal term to constrain the drift of each state. 
However, those methods suffer from two key drawbacks: (i) they treat safe data merely as a soft regularizer, which provides only weak control and makes it difficult to balance safety with downstream utility; and 
(ii) they typically select safe samples randomly, overlooking a fact that more relevant safe data can provide stronger corrective signals against harmful fine-tuning (as shown in Section~\ref{subsec:data_selection}).
This motivates the need for a more principled defense and data selection method to robustly withstand harmful fine-tuning.

To address these challenges, we propose \textbf{SPARD}, a novel framework that defends aligned LLMs against harmful fine-tuning attacks by combining \textbf{S}afety-\textbf{P}rojected \textbf{A}lternating optimization with \textbf{R}elevance-\textbf{D}iversity aware data selection. 
Figure~\ref{fig:SPARD_illustration} illustrates the overall procedure of SPARD. 
As shown, SPARD consists of two complementary components.
First, we study the safety-constrained fine-tuning problem, and introduce \emph{Safety-Projected Alternating Gradient} (SPAG), an optimization strategy that alternates between utility-driven updates on the fine-tuning data and explicit safety projections onto a constraint set defined by safe data.
Unlike penalty-based approaches, SPAG enforces feasibility in a closed form, ensuring that safety alignment is preserved throughout training.
Second, we recognize that the effectiveness of the safety projection critically depends on the choice of safe data.
That is, not all safe samples are equally informative: samples that align closely with the downstream task could provide stronger corrective signals than others.
To this end, we develop a \emph{Relevance–Diversity Determinantal Point Process (DPP)} that selects a compact subset of safe data to balance the task relevance and behavioral diversity, ensuring broad and effective coverage against harmful attacks.
Together, those two components yield a principled defense framework that maintains downstream utility while robustly constraining unsafe behaviors.

We conduct experiments on GSM8K~\citep{cobbe2021training} and OpenbookQA~\citep{mihaylov2018can} with four harmful finetuning attacks to evaluate both the utility and safety of SPARD. 
Empirical results show that SPARD can effectively mitigate harmful behaviors, achieving the lowest average Attack Success Rate (ASR) with high downstream accuracy. 
Moreover, SPARD significantly outperforms SafeInstr on average in ASR, showing the effectiveness of SPAG and Relevance–Diversity DPP.

Our contributions are summarized as follows: 
\begin{enumerate*}[(i), series = tobecont, itemjoin = \quad]
\item We propose SPARD, a novel defense framework that integrates safety-projected optimization with relevance–diversity–aware data selection to robustly defend aligned LLMs against harmful fine-tuning.  
\item We introduce SPAG, \textit{a principled optimization method} solving the novel safety-constrained fine-tuning problems that alternates between utility updates and explicit safety projections, and \textit{a Relevance–Diversity DPP} to select compact, task-aligned, and diverse safety subsets.  
\item Through extensive experiments on two aligned LLMs, multiple downstream tasks, and diverse attack datasets, we show that SPARD consistently outperforms existing defense methods in reducing the attack success rates while maintaining the utility.  
\end{enumerate*}



\section{Related Works}

\paragraph{LLM Safety and Alignment.}
Ensuring that large language models (LLMs) behave in a safe and helpful manner is a central challenge in AI research~\citep{yao2024survey}. 
Recent foundation models such as LLaMA~\citep{llama3-2, touvron2023llama2} and Qwen~\citep{yang2024qwen} have been aligned with safety guardrails to reject harmful instructions and follow user intent more reliably.
A common paradigm for alignment is preference-based learning, most notably Reinforcement Learning from Human Feedback~\citep{ouyang2022training, ziegler2019fine, bai2022training, lin2025parm}, which optimizes models to maximize human-preferred responses. 
Subsequent work has proposed more efficient formulations, such as Direct Preference Optimization~\citep{rafailov2023direct}, reward-free methods like RRHF~\citep{yuan2023rrhf}, which reduce reliance on expensive reward models while maintaining alignment quality. 
However, a critical vulnerability remains: the safety alignment achieved through these expensive procedures is often brittle and can be easily compromised or erased through subsequent downstream fine-tuning~\citep{yang2023shadow, yi2024vulnerability, qi2024finetuning, lermen2023lora, zhan2023removing, hsiung2025your, li2025salora}, which motivates the need for robust defense mechanisms.

\paragraph{Defending Against Harmful Fine-tuning.} 
Harmful fine-tuning attacks~\citep{huang2024harmful, liu2023jailbreaking, zou2023universal, yuan2023gpt} compromise aligned models by poisoning training data with adversarial prompts that bypass safety guardrails.
To counter such risks, many defense strategies~\citep{huang2024booster, huang2024vaccine, liu2025targeted, chen2025vulnerability, lyu2024keeping, hsu2024safe, bianchi2023safety, huang2024lisa, yi2025gradient, jiang2026metadefense} have been proposed.
For example,
MetaDefense~\citep{jiang2026metadefense} defends against finetuning-based jailbreak attacks by training a single LLM to predict the harmfulness of both incoming queries before generation and partial responses during generation.
PTST~\citep{lyu2024keeping} avoids safety degradation by fine-tuning solely on task data and re-introducing safety prompts at inference time.
SafeLoRA~\citep{hsu2024safe} constrains harmful updates by projecting LoRA weights from selected layers into a safety-aligned subspace.
Other approaches leverage safe data as an implicit safeguard. For example, SafeInstr~\citep{bianchi2023safety} mixes a small fraction of safe examples into fine-tuning data, while Lisa~\citep{huang2024lisa} uses safe samples by a bi-state optimization with a proximal regularization term.
Although effective to some extent, these methods either rely on carefully tuned penalty weights or only weakly address the utility–safety tradeoff.
In contrast, our proposed SPAG provides an optimization-grounded solution by explicitly projecting the model back into the safe region. This adaptive projection automatically determines the correction size, removing the need for manual weight tuning while simultaneously preserving downstream utility.

\paragraph{Data Selection for LLMs.}
The quality and composition of training data are critical determinants of LLM performance~\citep{zhou2023lima, gadre2023datacomp}.
This has motivated a growing body of work on data selection~\citep{albalak2024survey}, which seeks to curate smaller yet more effective subsets from vast, noisy corpora.
Selection strategies span a broad spectrum: filtering based on perplexity or linguistic complexity~\citep{longpre2024pretrainer}, identifying core sets that approximate the full dataset’s training dynamics~\citep{sorscher2022beyond}, or leveraging embedding similarity to retrieve samples closer to the target distribution for task-specific fine-tuning~\citep{liu2021makes, xia2024less, hsiung2025your, liu2025pharmacist}.
However, relevance-based selection alone often leads to redundancy. To mitigate this, we propose a novel approach to achieve a relevance-diversity trade-off.

\textbf{Determinantal Point Processes} (DPPs)~\citep{macchi1975coincidence, kulesza2012determinantal, chen2018fast} provide a principled probabilistic framework for subset selection that naturally achieves diversity.
Given a candidate pool $\hX = \{\vx_1, \dots, \vx_n\}$ and a positive semidefinite kernel matrix $\vL \in \mathbb{R}^{n \times n}$ with the kernel $\vL_{ij} = \hK(\vx_i,\vx_j)$, where $\hK(\cdot,\cdot)$ is the kernel function encoding similarity.
DPP assigns probability to each subset $\hC \subseteq \hX$ as
\begin{align}
\bP(\hC) = \frac{\det(\vL_{\hC})}{\det(\vI + \vL)},
\label{eq:dpp}
\end{align}
where $\vL_{\hC}$ is the principal submatrix indexed by $\hC$, $\vI$ is the identity matrix, and $\det(\cdot)$ is the determinant of a matrix.
The denominator $\det(\vI + \vL)$ is a constant independent of the subset selection, thus can be ignored.
Intuitively, $\det(\vL_{\hC})$ corresponds to the squared volume spanned by the feature vectors of $\hC$, favoring subsets whose elements are both individually informative and mutually dissimilar.
Yet, traditional DPPs ignore task relevance. Our work closes this gap by introducing a \emph{Relevance–Diversity DPP}, which incorporates task-relevance quality scores directly into the DPP kernel.



\section{Methodology}
\label{sec:methodology}

\subsection{Safety-Projected Alternating Gradient (SPAG)}
\label{subsec:spag}

In the proposed SPARD method, we formulate fine-tuning as a \emph{safety-constrained optimization problem} to adapt the model to downstream data without compromising its safety alignment:
\begin{align}
    \min_{\vtheta}\; \hL(\hD_{\text{ft}}, \vtheta)\quad\text{s.t.}\quad \hL(\hD_{\text{safe}}, \vtheta)\le \tau,
\end{align}
where $\hD_{\text{ft}}$ is the fine-tuning dataset, $\hD_{\text{safe}}$ is the safety dataset, and $\tau$ is a predefined threshold. 
In practice, $\tau$ can be set by measuring the average safety loss of the pretrained LLM on $\hD_{\text{safe}}$.

A common approach~\citep{huang2024lisa, yi2025gradient, bianchi2023safety} to relax the constraint is to add it to the objective as a penalty term:
\(
\min_{\vtheta}\; \hL(\hD_{\text{ft}}, \vtheta) + \lambda\big(\hL(\hD_{\text{safe}}, \vtheta) - \tau\big),
\)
with a penalty parameter $\lambda \ge 0$ controlling the balance between utility and safety.   
However, this blending lacks explicit control on safety, since the constraint is only enforced indirectly through the objective. 

Instead of implicitly encouraging safety via a soft penalty, we directly enforce the constraint using a projection-based strategy.  
The key idea is to first perform a utility-driven update (using $\hD_{\text{ft}}$) and then project the updated parameters back into a region where the safety constraint is approximately satisfied.  
This alternating update scheme avoids the difficulty of choosing penalty weights, while providing a principled geometric correction that guarantees feasibility up to first order.  

Specifically, after a utility update
\(
\vtheta^{+}=\vtheta-\eta_{\text{ft}}\nabla \hL(\hD_{\text{ft}}, \vtheta),
\)
where $\eta_{\text{ft}}$ is the learning rate for the utility step and $\nabla \hL(\hD_{\text{ft}}, \vtheta)$ denotes the gradient,
we project $\vtheta^{+}$ back into a linearized safety region determined by the safety loss.   
Using a first-order Taylor expansion at $\vtheta^{+}$, we approximate $\hL(\hD_{\text{safe}}, \vtheta)$ as 
\(
    \hL(\hD_{\text{safe}}, \vtheta)\approx \hL(\hD_{\text{safe}}, \vtheta^{+})+\langle \vg_{\text{safe}},\,\vtheta-\vtheta^{+}\rangle,
\)
where $\vg_{\text{safe}} = \nabla \hL(\hD_{\text{safe}}, \vtheta^{+})$.  
This defines the half-space
\(
    \mathcal{C}^{+}=\{\vtheta:\; \hL(\hD_{\text{safe}}, \vtheta^{+})+\langle \vg_{\text{safe}},\,\vtheta-\vtheta^{+}\rangle\le \tau\},
\)
which is a local approximation of the feasible set around $\vtheta^{+}$.

The projection step seeks the point in $\mathcal{C}^{+}$ that is closest to $\vtheta^{+}$:
\begin{align}
    \min_{\vtheta} &\quad \|\vtheta-\vtheta^{+}\|^{2} \nonumber \\
     \text{s.t.} & \quad
    \hL(\hD_{\text{safe}},  \vtheta^{+})+\langle \vg_{\text{safe}},\,\vtheta-\vtheta^{+}\rangle\le \tau,
     \label{eq:projection_qp}
\end{align}
Introducing multipliers for
the constraint in Eq. (\ref{eq:projection_qp}), the KKT conditions~\citep{bertsekas1997nonlinear} give the projected solution as
\begin{align}
	\vtheta^{\text{new}} \! = \!
	\begin{cases}
		\vtheta^{+}, & \!\!\! \text{if } \hL(\hD_{\text{safe}}, \vtheta^{+})\le \tau,\\[6pt]
		\vtheta^{+}- \frac{\hL(\hD_{\text{safe}}, \vtheta^{+})-\tau}{\|\vg_{\text{safe}}\|^{2}} \vg_{\text{safe}}, & \!\!\! \text{otherwise.}
	\end{cases}
    \label{eq:projected_solution}
\end{align}

The detailed derivation is provided in Appendix \ref{appendix:derivation-of-spag}.
When the safety condition is violated, to stabilize training by avoiding arbitrarily large projections, we adopt the trust-region optimization strategy~\citep{schulman2015trust} to limit the step size as 
 \(
\alpha=\min\!\left(\frac{\hL(\hD_{\text{safe}}, \vtheta^{+})-\tau}{\|\vg_{\text{safe}}\|^{2}}, \; \eta_{\text{safe}}\right)
\), and then $\vtheta^{\text{new}}=\vtheta^{+}- \alpha \,\vg_{\text{safe}}$,
where $\eta_{\text{safe}}$ is a trust-region radius.
The trust-region constraint ensures the updated model $\vtheta^{\text{new}}$ stays within a ball centered at $\vtheta^{+}$ with radius $ \eta_{\text{safe}} \|\vg_\text{safe}\|$.
According to Eq.~\eqref{eq:projected_solution}, the update either keeps $\vtheta^{+}$ if already safe, or applies a corrective step along $\vg_{\text{safe}}$ with magnitude determined by projection and trust region.

\begin{algorithm}[!t]
\captionof{algorithm}{SPAG.}
\begin{algorithmic}[1]
    \REQUIRE Fine-tuning dataset $\hD_{\text{ft}}$, safety dataset $\hD_{\text{safe}}$, learning rate $\eta_{\text{ft}}$, threshold $\tau$, trust-region radius $\eta_{\text{safe}}>0$; parameters $\vtheta$;
    \WHILE{not converged}
        \STATE {\color{Gray2}\COMMENT{Utility fine-tuning}}
        \STATE Sample mini-batch $\hB_{\text{ft}} \subseteq \hD_{\text{ft}}$;
        \STATE $\vtheta^{+} \leftarrow \vtheta - \eta_{\text{ft}} \nabla \hL(\hB_{\text{ft}}, \vtheta)$;
        \STATE {\color{Gray2}\COMMENT{Safety projection}}
        \STATE Sample mini-batch $\hB_{\text{safe}} \subseteq \hD_{\text{safe}}$;
        \STATE $\ell_{\text{safe}} \leftarrow \hL(\hB_{\text{safe}}, \vtheta^{+})$;
        \IF{$\ell_{\text{safe}} \le \tau$}
            \STATE $\vtheta \leftarrow \vtheta^{+}$;
        \ELSE
            \STATE $\vg_{\text{safe}} \leftarrow \nabla \hL(\hB_{\text{safe}}, \vtheta^{+})$;
            \STATE $\alpha \leftarrow \min\!\Big(\dfrac{\ell_{\text{safe}} - \tau}{\|\vg_{\text{safe}}\|^{2}}, \; \eta_{\text{safe}}\Big)$;
            \STATE $\vtheta \leftarrow \vtheta^{+} - \alpha\, \vg_{\text{safe}}$;
        \ENDIF
    \ENDWHILE
    \OUTPUT Safety-aligned parameters $\vtheta$.
\end{algorithmic}
\label{alg:spag}
\end{algorithm}

The algorithm of SPAG is illustrated in Algorithm \ref{alg:spag}.
In summary, SPAG provides a simple yet principled mechanism for safety-constrained fine-tuning: it alternates between utility optimization and explicit safety projection.  
Geometrically, the method corrects each update by projecting onto a safety half-space, thereby directly solving the constraint rather than relying on soft penalties.  
Unlike penalty-based approaches, which require careful tuning of $\lambda$ and offer no guarantee of feasibility, SPAG yields a closed-form projection step that enforces the constraint up to the first order.  
This combination of interpretability, guaranteed feasibility, and hyperparameter-free correction makes SPAG both practical and robust for safety-aligned fine-tuning.

\subsection{Relevance- and Diversity-Aware Safety Data Selection}
\label{subsec:data_selection}


While SPAG provides a principled projection mechanism for enforcing safety constraints, its success fundamentally depends on the quality of the safety dataset $\hD_{\text{safe}}$.
Recent studies have also observed that relevant safety data can improve safety during training by
leveraging embedding similarity~\citep{hsiung2025your}, 
employing trained selectors~\citep{liu2025pharmacist}, or matching task styles and formats~\citep{eiras2024safely, xiao2025style}.
Here we claim that \emph{not all safety samples contribute equally}. 
That is, safety samples, which align well with the fine-tuning domain $\hD_{\text{ft}}$, could provide stronger corrective signals than other safety samples.  
Therefore, carefully selecting \emph{relevant} safety data becomes critical to ensuring that the projection step effectively constrains the model.

\begin{figure}
    \centering
    \includegraphics[width=0.9\linewidth]{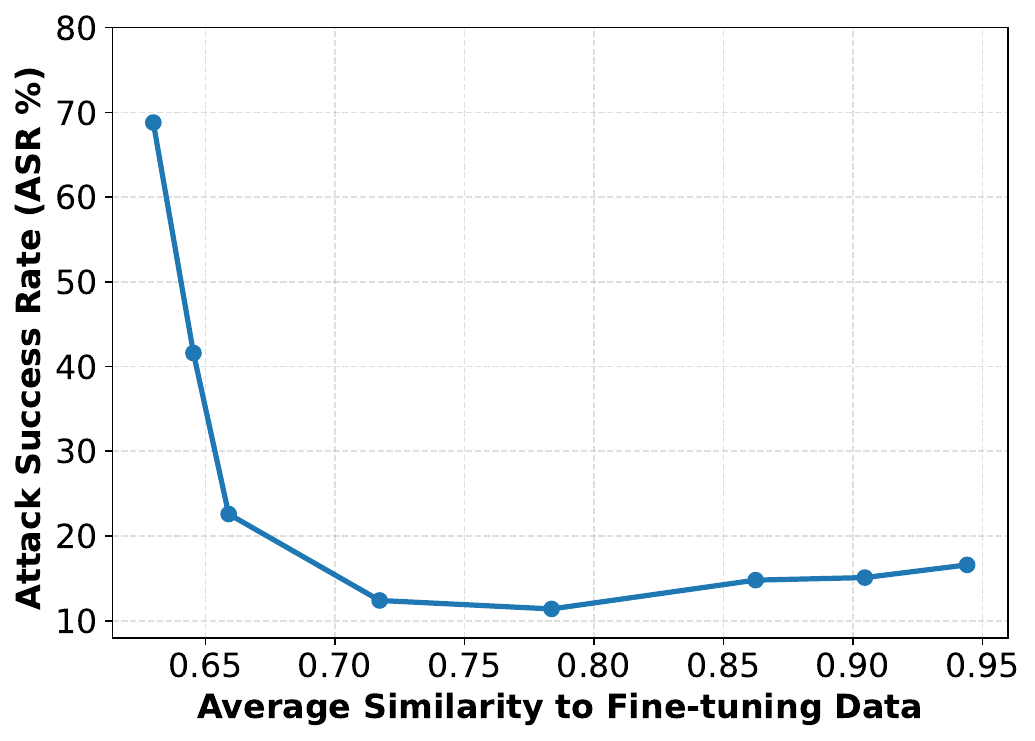}
    \caption{Attack success rate (ASR) with varying similarity levels of $\hD_{\text{safe}}$ to $\hD_{\text{ft}}$. The fine-tuning data are merged with BeaverTails attack data, and $\hD_{\text{safe}}$ are sampled from BeaverTails and LatHarmful defense data. }
    \vspace{-.1in}
    \label{fig:asr_vs_similarity}
\end{figure}

\textbf{Relevance Improves Safety.}
To better understand the role of relevance, we conduct an experiment, where the GSM8K training data are merged with 10\% BeaverTails attack data~\citep{ji2023beavertails} as $\hD_{\text{ft}}$, and safe samples are selected from BeaverTails and LatHarmful~\citep{sheshadri2024targeted} defense sets according to the similarity with $\hD_{\text{ft}}$.  
We define the similarity quality of a candidate $\vx_i \in \hD_{\text{safe}}$ as
\begin{align}
    q_i = \max_{\vx_z \in \hD_{\text{ft}}} \ssim(\vx_i, \vx_z),
    \label{similarity_fun}
\end{align}
where $\ssim(\cdot, \cdot)$ is cosine similarity in the embedding space.  
As shown in Figure~\ref{fig:asr_vs_similarity}, the attack success rate (ASR) decreases sharply as the average similarity of selected samples increases, dropping from $68.8\%$ at low similarity to $11.4\%$ at moderate-to-high similarity.  
This confirms that task-relevant safe samples provide stronger constraints and substantially enhance robustness. 
 
Notably, the curve also shows that ASR rises again (to $16.6\%$) when the selected samples are \emph{too similar} to the fine-tuning data (e.g., average similarity $\approx 0.94$).  
This counterintuitive phenomenon occurs because extreme similarity introduces redundancy: the selected safety samples cover only a narrow region of the risk space, leaving other harmful behaviors underrepresented.  
Consequently, the model may overfit to a small set of highly similar constraints, reducing the effectiveness of safety alignment.

\textbf{Relevance–Diversity DPP.}  
The above observation motivates a selection strategy that balances the \emph{relevance} and \emph{diversity}: the relevance ensures that the chosen samples provide strong and task-aligned safety signals, while the diversity ensures broad coverage of distinct harmful behaviors.
While \citet{hsiung2025your} introduces a metric for assessing subset diversity, they do not integrate this metric into the selection process, so their method cannot promote diversity during data selection.
To achieve this, we extend Determinantal Point Processes (DPPs)~\citep{macchi1975coincidence, kulesza2012determinantal, ye2023compositional}, which naturally promote diversity through determinant-based subset probabilities, by incorporating the relevance into the kernel design.  

Specifically, given relevance scores $q_i$ (defined in Eq. (\ref{similarity_fun})) for candidates $\vx_i \in \hD_{\text{safe}}$, We then construct the kernel as
\begin{align}
    \widehat{\hK}_{ij} = (q_i \cdot q_j)^\beta \cdot \hK(\vx_i, \vx_j),
\end{align}
where $\hK(\vx_i, \vx_j)$ captures the intrinsic similarity between two safety samples (e.g., the cosine similarity in an embedding space), and $\beta \ge 0$ controls the influence of the relevance.  
Intuitively, $(q_i \cdot q_j)^\beta$ acts as a multiplicative weight that increases the likelihood of including pairs of samples that are both highly relevant to the fine-tuning distribution (a sensitive analysis of $\beta$ is provided in Section~\ref{sec: analysis}).  
When $\beta=0$, the kernel reduces to the classical diversity-only DPP, while larger $\beta$ biases the distribution toward relevance-aware subsets.  

Let $\vL$ and $\widehat{\vL}$ be the kernel matrix corresponding to $\hK$ and $\widehat{\hK}$, respectively.
For any subset $\hC \subseteq \hD_{\text{safe}}$, according to Eq.~\eqref{eq:dpp}, the selection probability is given by
\begin{align}
    \bP(\hC) \propto \det(\widehat{\vL}_\hC) 
    = \prod_{\vx_i \in \hC} q_i^{2\beta} \cdot \det(\vL_{\hC}),
\end{align}
where $\vL_{\hC}$ denotes the kernel submatrix with indices in $\hC$.  
This decomposition makes the roles explicit: the factor $\prod q_i^{2\beta}$ rewards subsets containing highly relevant samples, while $\det(\vL_{\hC})$ enforces diversity among them.  
Thus, the relevance–diversity DPP jointly balances the task alignment and coverage, ensuring that the selected safety set is neither irrelevant nor redundant.  

\textbf{Efficient Greedy Selection.}  
Although DPPs define a principled probability distribution, exactly solving the maximum a posteriori (MAP) problem $\argmax_{\hC\subseteq \hD_{\text{safe}}} \det(\widehat{\vL}_\hC)$ is computationally expensive, requiring to calculate determinants of many submatrices.  
To scale the selection to large safety datasets, we adopt a greedy approximation that incrementally builds the subset by adding one sample at a time, each chosen to maximize the marginal gain in the determinant.  

Specifically, suppose we have already selected $\hC_{m-1}$ after $m-1$ steps.  
For a candidate $i \notin \hC_{m-1}$, the expanded kernel matrix is
\(
    \widehat{\vL}_{\hC_{m-1} \cup \{i\}} =
    \begin{pmatrix}
        \widehat{\vL}_{\hC_{m-1}} & \vv_i \\
        \vv_i^\top & \vL_{ii}
    \end{pmatrix},
\)
where $\vv_i$ contains kernel similarities between $\vx_i$ and the already-selected set $\hC_{m-1}$.  
By the Schur complement, the determinant after including $i$ can be factorized as
\begin{align}
\hspace{-0.2cm}\det(\widehat{\vL}_{\hC_{m-1} \cup \{i\}}) 
    = \det(\widehat{\vL}_{\hC_{m-1}}) \big(\vL_{ii} - \vv_i^\top \widehat{\vL}_{\hC_{m-1}}^{-1} \vv_i\big).
\label{Schur_complement}
\end{align}
The second term in the right-hand side of Eq. (\ref{Schur_complement}), known as the \emph{gain factor}, measures the additional volume contributed by $\vx_i$ that is not already spanned by $\hC_{m-1}$.  
Intuitively, it rewards candidates that are both individually relevant (large $\widehat{\vL}_{ii}$) and novel relative to the current set (small $\vv_i^\top \widehat{\vL}_{\hC_{m-1}}^{-1} \vv_i$).  

To compute the gain efficiently, we maintain the Cholesky decomposition $\widehat{\vL}_{\hC_{m-1}} = \vC \vC^\top$.  
Then,
\(
    \vv_i^\top \widehat{\vL}_{\hC_{m-1}}^{-1} \vv_i 
    = (\vC^{-1}\vv_i)^\top (\vC^{-1}\vv_i)
    = \|\vw_i\|^2,
\)
where $\vw_i$ is obtained by solving the triangular system $\vC \vw_i = \vv_i$.  
This avoids explicitly inverting $\widehat{\vL}_{\hC_{m-1}}$, reducing the complexity per iteration to $O(m)$ instead of cubic cost.  
At each step, we select
\(
    \vx_{i^\star} = \argmax_{\vx_i \in \hD_{\text{safe}} \setminus \hC_{m-1}} \Big(\widehat{\vL}_{ii} - \|\vw_i\|^2\Big),
\)
and add it into $\hC_{m-1}$ to obtain $\hC_{m}$. Additional details for computational cost are provided in Appendix~\ref{app:computational_complexity}.

\begin{table*}[!t]
    \centering
    \caption{Defense performance of Qwen-2.5-7B-Instruct on GSM8K under four harmful fine-tuning attacks. 
    Lower ASR/HS indicates stronger safety, while higher GSM8K accuracy reflects better utility. 
    Best results are in \textbf{bold}. GSM8K accuracy is averaged over all four attacks.}
    \label{tab:qwen2.5-gsm8k}
    \resizebox{0.97\textwidth}{!}{
    \begin{NiceTabular}{lcccccccccc|c}
    \toprule
          & \multicolumn{2}{c}{Beavertails} & \multicolumn{2}{c}{I-BeaverTails} & \multicolumn{2}{c}{LatHarmful} & \multicolumn{2}{c}{Q-LatHarmful} & \multicolumn{2}{c}{\textbf{Average}} &  {GSM8K} \\
         \cmidrule(lr){2-3} \cmidrule(lr){4-5} \cmidrule(lr){6-7} \cmidrule(lr){8-9} \cmidrule(lr){10-11} \cmidrule(lr){12-12}
         & ASR & HS & ASR & HS & ASR & HS & ASR & HS & ASR & HS & Accuracy \\
        \midrule
        Qwen-2.5-7B-Instruct & 25.80 & 1.71 & 36.06 & 1.97 & 7.88 & 1.23 & 6.52 & 1.22 & 19.02 & 1.53 & 77.71 \\ 
        \midrule
        SFT & 83.60 & 3.92 & 79.45 & 3.78 & 91.92 & 4.60 & 96.74 & 4.80 & 87.93 & 4.28 & \textbf{86.77} \\
        PTST~\citep{lyu2024keeping} & 62.40 & 3.03 & 70.23 & 3.29 & 82.42 & 4.13 & 83.50 & 4.11 & 74.64 & 3.64 & 85.06 \\
        SafeInstr~\citep{bianchi2023safety} & 65.60 & 3.21 & 66.25 & 3.18 & 76.77 & 3.99 & 85.74 & 4.34 & 73.59 & 3.68 & 86.28 \\
        Lisa~\citep{huang2024lisa} & 24.40 & 1.73 & 35.64 & 2.00 & \textbf{7.88} & \textbf{1.26} & 8.55 & \textbf{1.25} & 19.12 & 1.56 & 78.45 \\
        SafeGrad~\citep{yi2025gradient} & 32.00 & 1.97 & 41.72 & 2.22 & 20.00 & 1.75 & 27.00 & 1.77 & 30.18 & 1.93 & 85.71 \\
        \rowcolor{Gray}
        SPARD & \textbf{10.00} & \textbf{1.32} & \textbf{12.58} &  \textbf{1.36} & \textbf{7.88} & 1.31 &  \textbf{7.33} & 	1.29 & \textbf{9.45} & \textbf{1.32} & 85.77 \\
    \bottomrule
    \end{NiceTabular}
    }
\end{table*}

\begin{table*}[!t]
    \centering
    \caption{Defense performance of LLaMA-3.2-3B-Instruct on GSM8K under four harmful fine-tuning attacks.
    Best results (lowest ASR/HS and highest GSM8K accuracy) are in \textbf{bold}. GSM8K accuracy is averaged over all four attacks. }
    \label{tab:llama3b-gsm8k}
    \resizebox{0.97\textwidth}{!}{
    \begin{NiceTabular}{lcccccccccc|c}
    \toprule
          & \multicolumn{2}{c}{Beavertails} & \multicolumn{2}{c}{I-BeaverTails} & \multicolumn{2}{c}{LatHarmful} & \multicolumn{2}{c}{Q-LatHarmful} & \multicolumn{2}{c}{\textbf{Average}} & {GSM8K} \\
         \cmidrule(lr){2-3} \cmidrule(lr){4-5} \cmidrule(lr){6-7} \cmidrule(lr){8-9} \cmidrule(lr){10-11} \cmidrule(lr){12-12}
         & ASR & HS & ASR & HS & ASR & HS & ASR & HS & ASR & HS &  Accuracy \\
         \midrule
        LLaMA-3.2-3B-Instruct & 41.80 & 2.12 & 52.20 & 2.44 & 16.36 & 1.55 & 12.02 & 1.40 & 30.60 & 1.88 & 62.32 \\ 
        \midrule
        SFT & 87.20 & 4.03 & 79.45 & 3.66 & 98.99 & 4.91 & 99.80 & 4.95 & 91.36 & 4.39 & 72.27 \\
        PTST~\citep{lyu2024keeping} & 58.80 & 2.87 & 64.99 & 3.04 & 97.78 & 4.82 & 96.33 & 4.73 & 79.48 & 3.87 & \textbf{73.75} \\
        SafeInstr~\citep{bianchi2023safety} & 76.20 & 3.61 & 72.54 & 3.46 & 90.10 & 4.56 & 89.21 & 4.52 & 82.01 & 4.04 & 72.21 \\
        Lisa~\citep{huang2024lisa} & 32.20 & 1.95 & 45.70 & 2.32 & \textbf{9.90} & \textbf{1.36} & \textbf{8.96} & \textbf{1.29} & 24.19 & 1.73 & 65.03 \\
        SafeGrad~\citep{yi2025gradient} & 50.80 & 2.45 & 59.12 & 2.77 & 78.79 & 4.06 & 98.37 & 4.86 & 71.77 & 3.54 & 64.29 \\
        \rowcolor{Gray} 
        SPARD & \textbf{13.20} & \textbf{1.33} & \textbf{14.68} & \textbf{1.42} & 11.31 & 1.43 &  9.16 & 1.35 & \textbf{12.09} & \textbf{1.38} & 71.23 \\
    \bottomrule
    \end{NiceTabular}
    }
\end{table*}

\section{Experiments}
\label{sec:expt}

\subsection{Experimental Setup}
\label{sec:expt:setup}

\textbf{Datasets.}  
\emph{Safety corpora.} We use four datasets containing harmful or jailbreak-style prompts with both harmful and safe responses:  
(i) {BeaverTails}~\citep{ji2023beavertails}, a collection of safety-related QA pairs with helpfulness and harmlessness annotations;  
(ii) {I-BeaverTails}, constructed by converting BeaverTails questions into instructions using GPT-4o-mini~\citep{hurst2024gpt} following \citet{bianchi2023safety};  
(iii) {LatHarmful}~\citep{sheshadri2024targeted}, consisting of 5k instructions with paired harmful and harmless completions; and  
(iv) {Q-LatHarmful}, obtained by converting LatHarmful instructions into QA pairs with GPT-4o-mini.  
Each dataset is split 90\%/10\% into training and testing. From these corpora, we use \emph{harmful queries with safe responses} from the training splits to build \textbf{GeneralSafe}, the candidate pool for selecting safety data. In contrast, to simulate harmful fine-tuning, we use \emph{harmful queries with harmful responses} from the training splits to inject malicious samples into downstream utility tasks.
Following~\citet{huang2024lisa, hsu2024safe}, the number of injected harmful samples is fixed to 10\% of the original utility task training size, ensuring a consistent attack intensity across tasks.   

\emph{Utility tasks.} For downstream performance, we evaluate on  
(i) {GSM8K}~\citep{cobbe2021training}, a benchmark of grade-school math word problems, augmented with MetaMath~\citep{yu2023metamath} for broader coverage; and  
(ii) {OpenBookQA}~\citep{mihaylov2018can}, a science QA dataset requiring factual reasoning.    

\textbf{Evaluation Metrics.} 
Following \citet{hsu2024safe}, we evaluate methods on two dimensions: \textit{Safety} and \textit{Utility}. 
For safety, we adopt the protocol of \citet{qi2024finetuning}, using GPT-4o-mini to judge responses under OpenAI's 11 harmful content categories. 
Each response receives a Harmfulness Score (HS) from 1 (safest) to 5 (most harmful), and we report the Attack Success Rate (ASR), the proportion of responses with HS $>2$. 
For utility, we measure accuracy on downstream tasks (GSM8K or OpenBookQA), reflecting utility performance under when safety defense is enforced.

\textbf{Baselines.}
We compare SPARD against a range of baselines using two safe pre-trained models, Qwen-2.5-7B-Instruct~\citep{yang2024qwen} and LLaMA-3.2-3B-Instruct~\citep{llama3-2}. 
Specifically, we consider:
\begin{enumerate*}[(i), series = tobecont, itemjoin = \quad]
\item SFT, which is a standard fine-tuning baseline where the model is trained exclusively on the target task data without any explicit safety measures.
\item PTST~\citep{lyu2024keeping}, which fine-tunes the model without safety instructions but prepends them back to inputs at inference time. 
\item SafeInstr~\citep{bianchi2023safety}, which randomly mixes a small fraction (3\%) of safe samples into the fine-tuning dataset.
\item Lisa~\citep{huang2024lisa}, which bi-state learning finetuning samples and safe samples with a proximal term to constrain the safety degradation of each state. 
\item SafeGrad~\citep{yi2025gradient}, which detects conflict safety/alignment gradients and projects out the harmful component.
\end{enumerate*}

\begin{table*}[!t]
    \centering
    \caption{Defense performance of Qwen-2.5-7B-Instruct on OpenBookQA under four in-distribution harmful fine-tuning attacks. 
    Lower ASR/HS indicates stronger safety; higher accuracy indicates better utility. 
    OpenBookQA accuracy is averaged over the four attacks. Best results are in \textbf{bold}.}
    \label{tab:qwen2.5-openbookqa}
    \resizebox{0.97\textwidth}{!}{
    \begin{NiceTabular}{lcccccccccc|c}
    \toprule
          & \multicolumn{2}{c}{Beavertails} & \multicolumn{2}{c}{I-BeaverTails} & \multicolumn{2}{c}{LatHarmful} & \multicolumn{2}{c}{Q-LatHarmful} & \multicolumn{2}{c}{\textbf{Average}} & {OpenbookQA} \\
         \cmidrule(lr){2-3} \cmidrule(lr){4-5} \cmidrule(lr){6-7} \cmidrule(lr){8-9} \cmidrule(lr){10-11} \cmidrule(lr){12-12} 
         & ASR & HS & ASR & HS & ASR & HS & ASR & HS & ASR & HS & Accuracy \\
         \midrule
        Qwen-2.5-7B-Instruct & 25.80 & 1.71 & 36.06 & 1.97 & 7.88 & 1.23 & 6.52 & 1.22 & 19.02 & 1.53 & 77.60 \\ 
        \midrule
        SFT & 50.20 & 2.70 & 56.39 & 2.81 & 25.66 & 1.96 & 28.92 & 2.05 & 40.29 & 2.38 & \textbf{83.70} \\
        PTST~\citep{lyu2024keeping} & 37.20 & 2.09 & 56.39 & 2.81 & 10.71 & 1.35 & 16.90 & 1.59 & 30.30 & 1.96 & 83.25 \\
        SafeInstr~\citep{bianchi2023safety} & 51.40 & 2.68 & 59.75 & 2.95 & 26.06 & 1.95 & 29.53 & 2.11 & 41.69 & 2.42 & 84.15 \\
        Lisa~\citep{huang2024lisa} & 26.00 & 1.74 & 34.80 & 1.93 & \textbf{7.07} & \textbf{1.22} & \textbf{7.33} & \textbf{1.24} & 18.80 & 1.53 & 78.90 \\
        SafeGrad~\citep{yi2025gradient} & 28.80 & 1.80 & 36.29 & 2.01 & 7.68 & 1.24 & 9.78 & 1.32 & 20.63 & 1.59 & 83.30 \\
        \rowcolor{Gray}
        SPARD & \textbf{21.80} & \textbf{1.65} & \textbf{20.13} & \textbf{1.57} & 8.28 & 1.31 & 7.94 & 1.27 & \textbf{14.54} & \textbf{1.45} & 83.25 \\
    \bottomrule
    \end{NiceTabular}
    }
    \vspace{-.1in}
\end{table*}

\begin{figure*}[!t]
\begin{minipage}[b]{0.32\linewidth}
\centering
\includegraphics[width=0.99\linewidth]{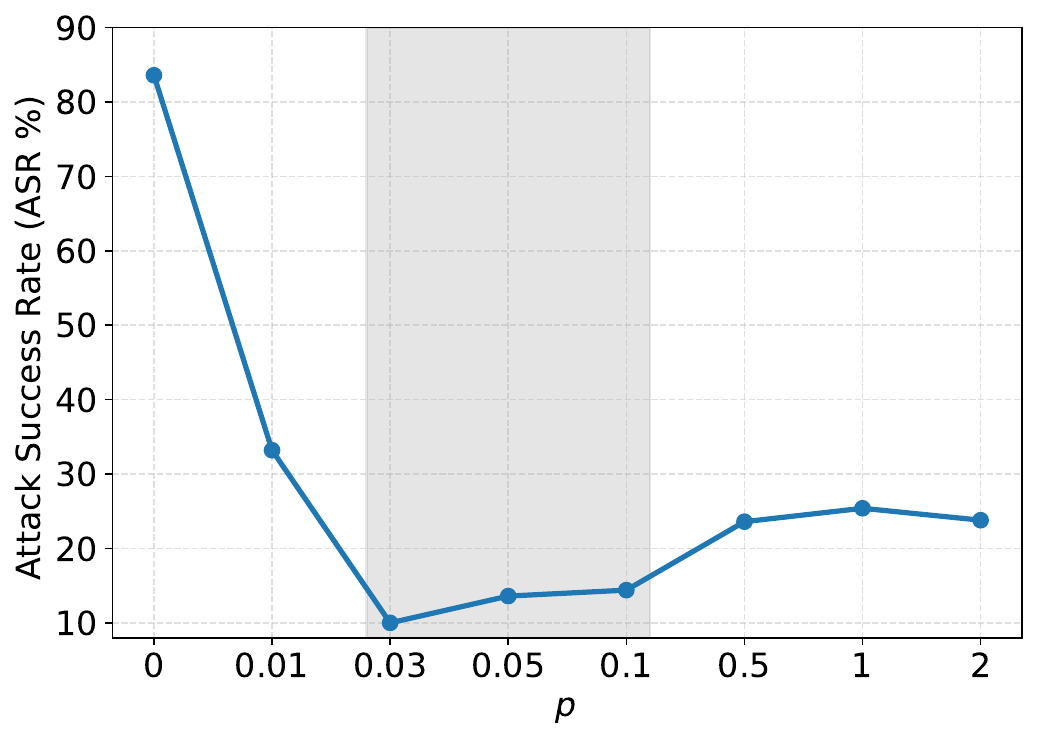}
\vspace{-.2in}
\caption{Effects of $p$.}
\label{fig:safe_ratio}
\end{minipage}  
\hfill
\begin{minipage}[b]{0.32\linewidth}
\centering
\includegraphics[width=0.99\linewidth]{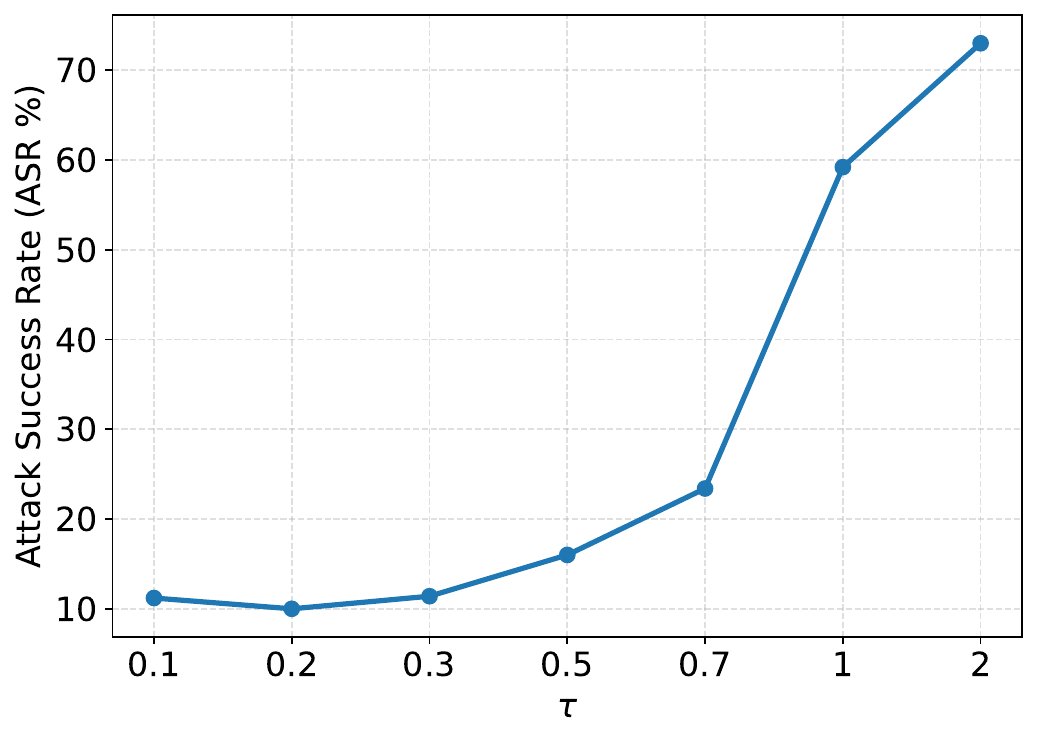}
\vspace{-.2in}
\caption{Effects of $\tau$.}
\label{fig:tau}
\end{minipage} 
\hfill
\begin{minipage}[b]{0.32\linewidth}
\centering
\includegraphics[width=0.99\linewidth]{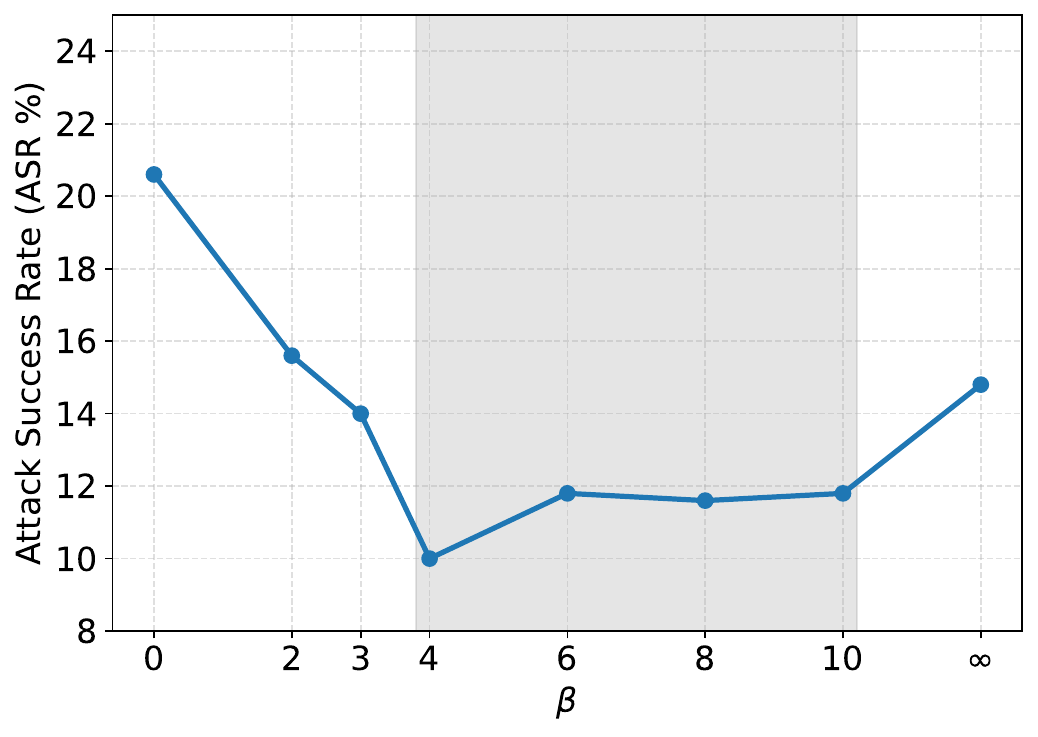}
\vspace{-.2in}
\caption{Effects of $\beta$.}
\label{fig:beta}
\end{minipage} 
\vskip -.2in
\end{figure*}

\paragraph{Implementation Details}
We employ LoRA~\citep{hu2022lora} for parameter-efficient fine-tuning, with a rank $r=32$ and an alpha of $4$. 
All models are trained using the AdamW optimizer~\citep{loshchilov2017decoupled}. 
We set the learning rate to $5 \times 10^{-5}$ for both Qwen-2.5-7B-Instruct and LLaMA-3.2-3B-Instruct. 
The models are fine-tuned for 10 epochs on GSM8K and 3 epochs on OpenBookQA.
For our SPAG algorithm, the safety mini-batch $\hB_\text{safe}$ is sampled with a batch size of 1. 
The trust region radius $\eta_\text{safe}$ is set equal to the fine-tuning learning rate $\eta_\text{ft}$. 
The hyperparameters $\tau$, $\delta$, and $\epsilon$ are chosen as $0.2$, $0.1$, and $1 \times 10^{-8}$ for all experiments.
For our Relevance-Diversity DPP data selection, sample embeddings are generated by taking the average of the final layer's hidden states from the pretrained model (i.e., Qwen-2.5-7B-Instruct and LLaMA-3.2-3B-Instruct). 
Based on these embeddings, we select $\hD_\text{safe}$ from the \textbf{GeneralSafe} pool, with the size fixed to $3\%$ (i.e., $p=0.03$) of the fine-tuning data. 
The relevance exponent is set to $\beta=4$.

\subsection{Main Result}
\label{sec:expt:main}

\textbf{Robustness to Different Attacks.}
Table~\ref{tab:qwen2.5-gsm8k} presents results on GSM8K under four harmful fine-tuning attacks. 
As can be seen, SPARD simultaneously achieves the lowest ASR/HS while preserving competitive GSM8K accuracy, offering the strongest balance between safety and utility among all methods.

Specifically, compared with SFT and PTST, SPARD achieves substantial gains: average ASR is reduced by over $65\%$ and HS by $2.32$ points, while accuracy remains competitive. This shows that explicit safety projection is far more effective than standard fine-tuning or inference-time prompting.  
Compared with SafeInstr, which randomly mixes safe samples into fine-tuning, SPARD consistently achieves lower ASR/HS, highlighting the necessity of principled relevance–diversity selection over naive random selection.  
Compared with Lisa, a strong optimization-based baseline, SPARD achieves consistently better safety on all datasets, reducing ASR by $9.67\%$, lowering HS by $0.24$, and significantly improving GSM8K accuracy by $7.32\%$.
Compared with SafeGrad, SPARD achieves a substantially lower ASR of $20.73\%$ while having a slightly better downstream performance.
This demonstrates that the key designs of SPARD—SPAG safety projection and relevance–diversity DPP selection—are crucial for constraining harmful behaviors while preserving downstream task performance.

\textbf{Generalization Across Architectures (LLaMA).}
Table~\ref{tab:llama3b-gsm8k} presents results on LLaMA-3.2-3B-Instruct. 
We observe that the overall safety degradation is more severe on LLaMA than on Qwen, as SFT and PTST both yield very high ASR/HS despite maintaining task accuracy. 
SPARD, however, remains effective across model families, achieving the lowest ASR ($12.09\%$) and HS ($1.38$) while keeping accuracy competitive ($71.23\%$).  
Compared with SFT and PTST, SPARD lowers ASR by over $67\%$, confirming that its safety projection generalizes across backbones. 
Against SafeInstr, which suffers from high ASR/HS, SPARD shows the value of relevance–diversity selection over naive random mixing.  
Relative to Lisa, SPARD further improves both safety and utility ($-12.10\%$ ASR, $-0.35$ HS, $+6.20\%$ accuracy). 
SPARD also outperforms SafeGrad, reducing ASR by over $59\%$ while achieving a substantial $6.94\%$ improvement in accuracy.
Together, these results demonstrate that SPAG optimization and DPP-based selection enhance robustness consistently across architectures.

\textbf{Generalization to OpenBookQA.} 
Table~\ref{tab:qwen2.5-openbookqa} reports results on OpenBookQA under four harmful fine-tuning attacks. 
SPARD achieves the lowest ASR ($14.54\%$) and HS ($1.45$) while preserving strong task accuracy ($83.25\%$), confirming that its effectiveness extends beyond math reasoning tasks.  
Compared with SFT and PTST, SPARD reduces ASR by more than $15.3\%$ on average, showing that explicit safety projection remains effective for science QA.  
Relative to SafeInstr, which again suffers from high ASR/HS, SPARD demonstrates the importance of relevance–diversity selection over naive data mixing. 
Compared with Lisa, SPARD achieves lower ASR/HS ($-4.14\%$/$-0.05$) and higher accuracy ($+4.05\%$), reinforcing that its joint use of SPAG optimization and DPP-based selection improves both safety and utility. 
Finally, SPARD outperforms SafeGrad with an improvement of $5\%$ on ASR, validating the effectiveness of SPAG's safety projection.
These results highlight that SPARD generalizes beyond GSM8K to diverse downstream reasoning tasks, maintaining robustness across domains.

\begin{table*}[!t]
\centering
\caption{Effect of the trust-region radius $\eta_{\text{safe}}$ of Qwen-2.5-7B-Instruct on GSM8K.}
\resizebox{0.9\textwidth}{!}{%
\begin{NiceTabular}{lcccccc}
\toprule
Method & BeaverTails & I\text{-}BeaverTails & LatHarmful & Q\text{-}LatHarmful & \textbf{Avg. ASR}  & GSM8K Acc \\
\midrule
SPARD (w/o $\eta_{\text{safe}}$) & 8.20 & 10.69 & 0.00 & 1.22 & 5.03 & 81.92 \\
SPARD & 10.00 & 12.58 & 7.88 &  7.33  & 9.45  & 85.77 \\
\bottomrule
\end{NiceTabular}
}
\vspace{-.05in}
\label{tbl:abl_eta_safe}
\end{table*}

\begin{table*}[t]
    \centering
    \caption{Effect of Relevance-Diversity DPP under different harmful fine-tuning attacks. 
    Best results are in \textbf{bold}.}
    \label{tab:ablation_dpp}
    \resizebox{0.97\textwidth}{!}{
    \begin{NiceTabular}{lcccccccccc|c}
    \toprule
          & \multicolumn{2}{c}{Beavertails} & \multicolumn{2}{c}{I-BeaverTails} & \multicolumn{2}{c}{LatHarmful} & \multicolumn{2}{c}{Q-LatHarmful} & \multicolumn{2}{c}{\textbf{Average}} &  {GSM8K} \\
         \cmidrule(lr){2-3} \cmidrule(lr){4-5} \cmidrule(lr){6-7} \cmidrule(lr){8-9} \cmidrule(lr){10-11} \cmidrule(lr){12-12}
         & ASR & HS & ASR & HS & ASR & HS & ASR & HS & ASR & HS & Accuracy \\
        \midrule
        Lisa~\citep{huang2024lisa} & 24.40 & 1.73 & 35.64 & 2.00 & 7.88 & 1.26 & 8.55 & 1.25 & 19.12 & 1.56 & 78.45 \\
        \midrule
        SPAG \textit{w/} Random & 15.80 & 1.47 & 22.64 & 1.70 & 13.33 & 1.52 & 11.61 & 1.46 & 15.85 & 1.54 & 85.06 \\
        SPAG \textit{w/} Max Quality & 16.60 & 1.53 & 24.32 & 1.71 & 16.16 & 1.63 & 8.96 & 1.34 & 16.51 & 1.55 & 85.69 \\
        \rowcolor{Gray}
        SPARD & \textbf{10.00} & \textbf{1.32} &	\textbf{12.58} &  \textbf{1.36} & \textbf{7.88}& \textbf{1.31} &  \textbf{7.33} & 	\textbf{1.29} & \textbf{9.45} & \textbf{1.32} & \textbf{85.77} \\
    \bottomrule
    \end{NiceTabular}
    }
    \vspace{-.1in}
\end{table*}

\subsection{Analysis}
\label{sec: analysis}

\textbf{Effects of safe sample ratio.}  
Figure~\ref{fig:safe_ratio} shows the impact of varying the ratio $p$ of safe samples added to the GSM8K finetuning data using Qwen-2.5-7B-Instruct under the BeaverTails attack. 
When $p=0$ (i.e., no safe samples are added), the model is highly vulnerable with ASR above $80\%$. As $p$ increases, ASR drops sharply and reaches the lowest point around $p \in [0.03, 0.05]$. 
Beyond this range, further increasing $p$ leads to diminishing returns and even slight degradation due to the inclusion of redundant or less relevant samples. 
This indicates that a small but carefully chosen proportion $p\in [0.03, 0,1]$ of safe samples is sufficient to provide strong safety guarantees without overwhelming the fine-tuning objective.  

\textbf{Effects of $\tau$.}  
We study the effects of the safety threshold $\tau$ on GSM8K using Qwen-2.5-7B-Instruct under the BeaverTails attack. 
As shown in Figure~\ref{fig:tau}, small values of $\tau$ enforce strict safety constraints, effectively suppressing ASR, but overly conservative thresholds ($\tau > 0.5$) begin to harm the balance and allow ASR to rise again. 
In practice, we can set the $\tau$ by referencing the average loss of the aligned LLM on the safety benchmark.

\textbf{Effects of $\beta$.}  
To analyze the effect of relevance exponent $\beta$, we conduct experiments with the BeaverTails attack on GSM8K using Qwen-2.5-7B-Instruct.
Figure~\ref{fig:beta} analyzes the relevance exponent $\beta$, which balances the weight between relevance and diversity in the DPP kernel.  
SPARD is relatively robust to a wide range of moderate values ($\beta \in  [4,10]$), achieving the lowest ASR, while very small $\beta$ underemphasizes relevance and very large $\beta$ collapses diversity, both leading to weaker defenses. 
These results confirm that both relevance and diversity should be considered in the data selection process.

\noindent \textbf{Effects of} $\eta_{\text{safe}}.$
We conduct an ablation removing the trust-region limit $\eta_{\text{safe}}$. 
Table~\ref{tbl:abl_eta_safe} reports the ASR under multiple attacks and the average downstream GSM8K accuracy. 
Removing $\eta_{\text{safe}}$ results in more aggressive projection updates: safety improves for some attacks, but downstream utility degrades substantially. These results highlight that $\eta_{\text{safe}}$ plays a crucial role in stabilizing the projection step, achieving a more balanced trade-off between strong safety and good downstream performance.

\begin{table}[!t]
    \centering
    \caption{Defense performance with full fine-tuning on SmolLM2-1.7B-Instruct (GSM8K, BeaverTails attack). Best results are in \textbf{bold}.}
    \label{tab:full-ft}
    \begin{NiceTabular}{lccc}
        \toprule
        Method & Accuracy & ASR & HS \\
        \midrule
        SFT & \textbf{51.5} & 74.0 & 3.61 \\
        Lisa~\citep{huang2024lisa} & 49.4 & 69.0 & 3.40 \\
        SafeGrad~\citep{yi2025gradient} & 50.4 & 23.2 & 1.71 \\
        SPARD & 50.6 & \textbf{17.4} & \textbf{1.55} \\
        \bottomrule
    \end{NiceTabular}
    \vspace{-.2in}
\end{table}

\begin{figure*}[t]
	\centering
	\!\!\!
	\subfigure[Random.\label{fig:random-selection}]{\includegraphics[width=0.32\textwidth]{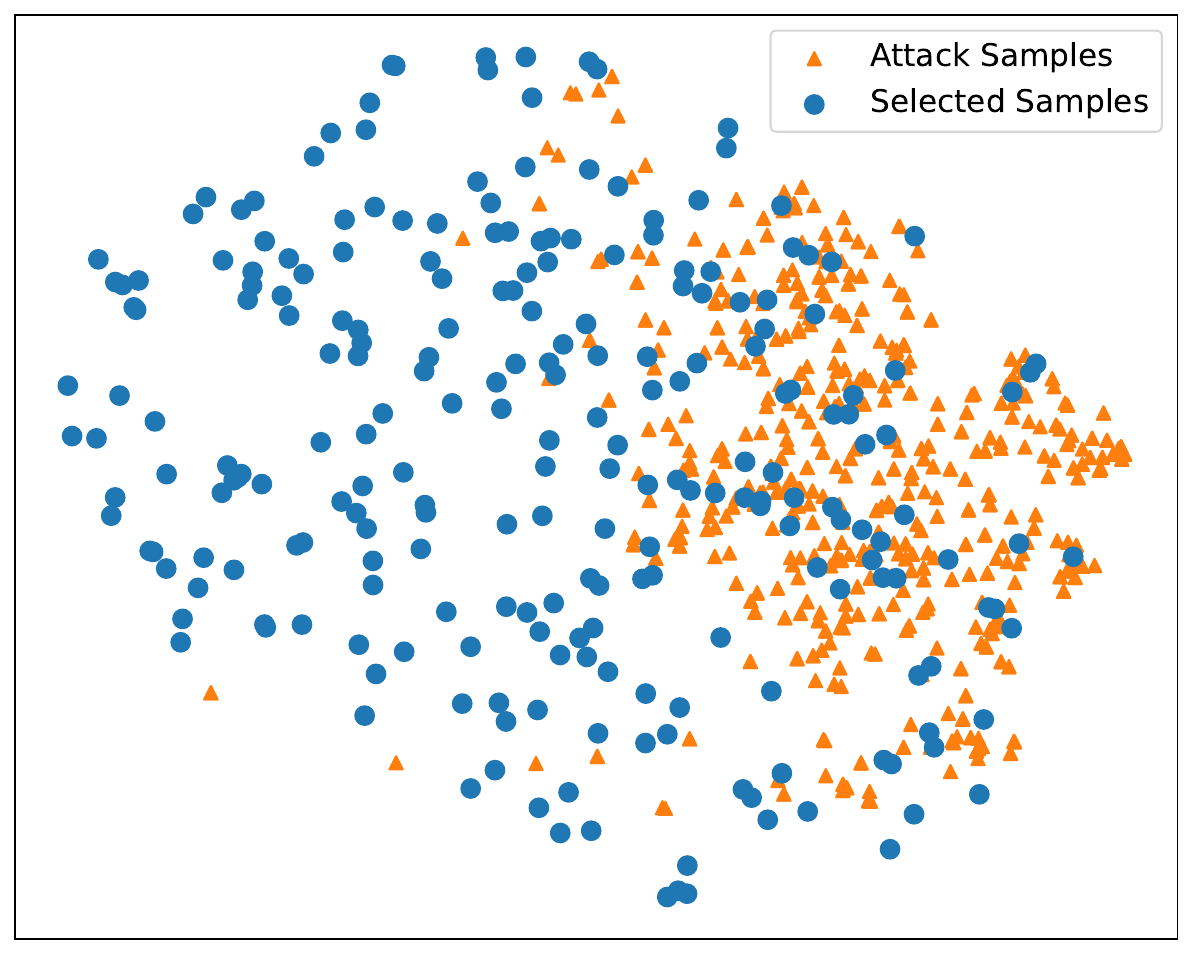}} 
	\subfigure[Max Quality (i.e., $\beta=+\infty$).\label{fig:cloest-selection}]{\includegraphics[width=0.32\textwidth]{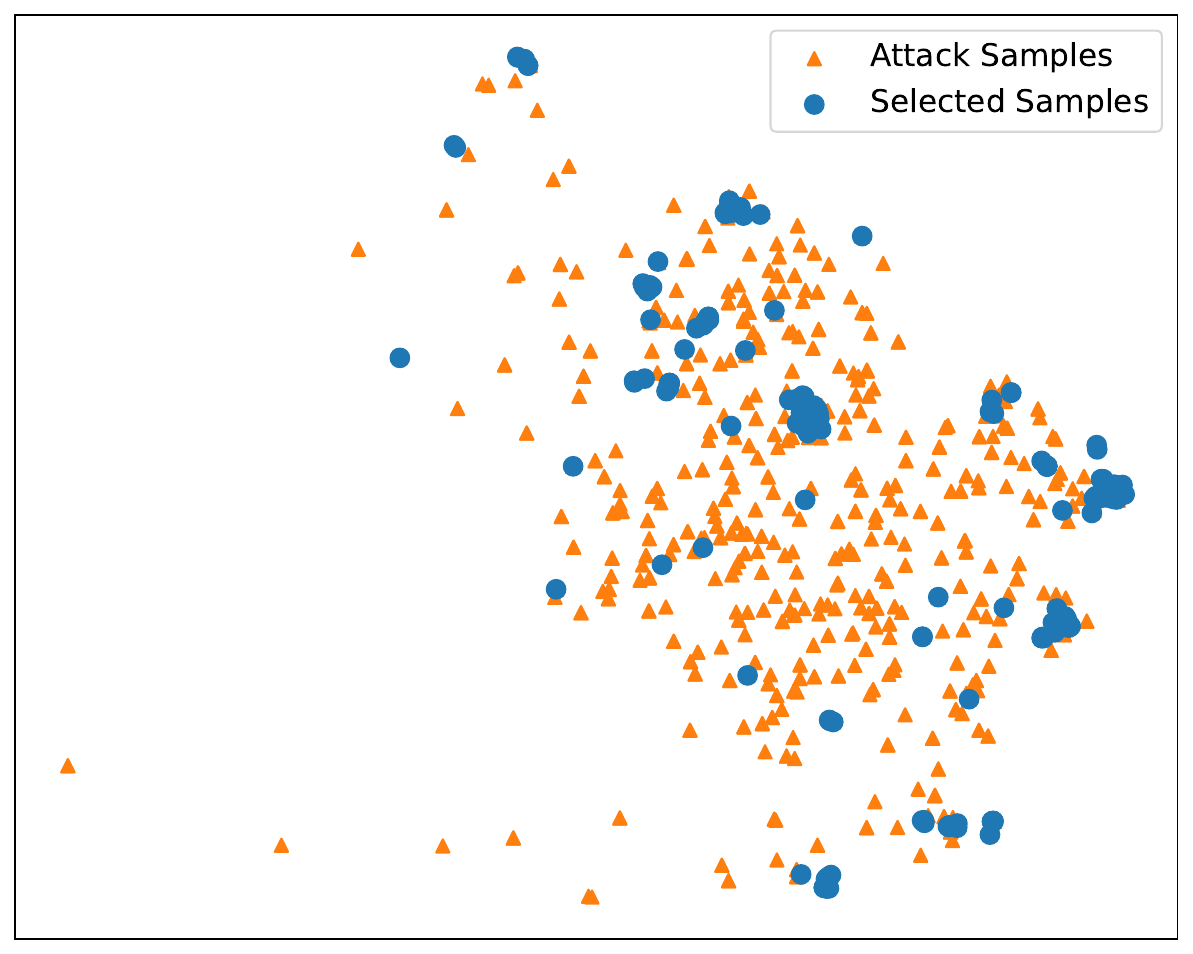}}
	\subfigure[SPARD.\label{fig:our-selection}]{\includegraphics[width=0.32\textwidth]{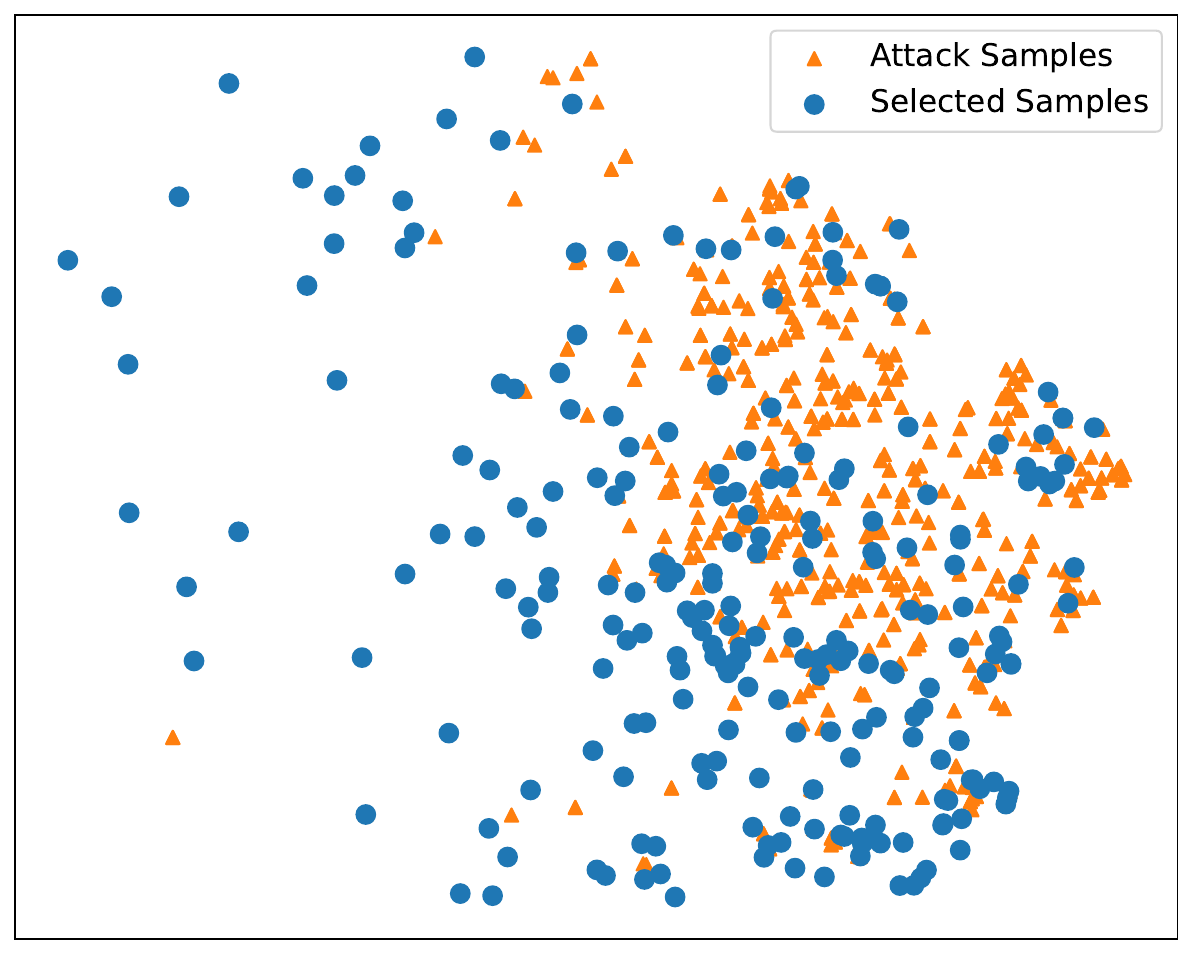}}\!\!\!
    \vspace{-.1in}
	\caption{Comparison of selected safe samples for GSM8K task under BeaverTails attack.}
	\label{fig:tsne-selection-comparison}
    \vspace{-.15in}
\end{figure*}

\textbf{Effect of Relevance-Diversity DPP.}
To study the effect of Relevance-Diversity DPP, we compare it with 
(i) SPAG \textit{w/} Random, which randomly selects samples from GeneralSafe as $\hD_\text{safe}$.
(ii) SPAG \textit{w/} Max Quality, which selects the samples with the highest quality score as $\hD_\text{safe}$.
As shown in Table~\ref{tab:ablation_dpp}, SPAG \textit{w/} Random surpasses previous SOTA (i.e., Lisa) with an average ASR reduction of $3.27\%$ and an GSM8K accuracy improvement of $6.61\%$, validating the effectiveness of SPAG safety projection.
Compared with all variants, SPARD has the best average safety and utility, achieving the lowest mean ASR/HS ($9.45\%$/$1.32$) and the highest GSM8K accuracy ($85.77\%$).
Specifically, SPARD outperforms SPAG \textit{w/} Random with a noticeable ASR and HS reduction of $6.40\%$ and $0.22$, showing that selecting relevant data can substantially improve safety.
Additionally, SPARD surpasses SPAG \textit{w/} Max Quality by a large margin of $7.06\%$ on average ASR, validating that diversity is equally crucial. 
By balancing both relevance and diversity, SPARD achieves broad coverage of safety constraints while remaining task-aligned, leading to superior robustness without sacrificing utility.
Moreover, as \citet{hsiung2025your} also explores similarity and diversity metrics in safety data curation, we provide further discussion on the similarities and differences, along with an empirical comparison of the two methods, in Appendix~\ref{app:additional_selection}.


\noindent \textbf{Full Fine-Tuning Without LoRA.}
SPAG's projection (Eq.~\eqref{eq:projected_solution}) operates on whatever parameters are being updated, independent of LoRA structure. To verify this, we evaluate SPARD with full fine-tuning on SmolLM2-1.7B-Instruct (GSM8K, BeaverTails attack).
As shown in Table~\ref{tab:full-ft}, SPARD achieves the lowest ASR ($17.4\%$) and HS ($1.55$) while maintaining competitive accuracy, confirming that the findings are consistent without LoRA.

\noindent \textbf{Visualization.}
Figure~\ref{fig:tsne-selection-comparison} shows the t-SNE visualization~\citep{van2008visualizing} of selected samples for the GSM8K task under BeaverTails attacks. 
As shown, randomly selected data cover diverse regions but are not necessarily aligned with the attacked distribution, leading to limited safety gains.
Moreover, a quality-only strategy (Max Quality) selects samples that cluster tightly around the attack distribution, but suffers from severe redundancy.
In contrast, SPAG achieves a balanced selection that aligns samples closely with the attacked distribution while maintaining diversity across different safety corpora, ensuring broad coverage without redundancy.
This suggests that our method is effective in selecting safe samples that are both relevant to the task and diverse (Tables \ref{tab:ablation_dpp}).

\begin{table}[t]
\centering
\caption{End-to-end wall-clock time (minutes) and relative overhead for fine-tuning Qwen-2.5-7B-Instruct on GSM8K with a single A800 GPU. Overhead is measured relative to vanilla SFT.}
\label{tab:overhead}
\begin{NiceTabular}{lcc}
\toprule
Method & Time (min) & Overhead vs.\ SFT \\
\midrule
SFT & 253.0 & - \\
PTST & 253.0 & +0\% \\
SafeInstr & 260.3 & +2.9\% \\
Lisa & 277.8 & +9.8\% \\
SafeGrad & 390.5 & +54.2\% \\
SPARD & 273.5 & +8.1\% \\
\bottomrule
\end{NiceTabular}
\vspace{-.25in}
\end{table}

\noindent \textbf{Computational Overhead.}
The preprocessing unique to SPARD is lightweight: on a single A800 GPU, embedding extraction takes 3.42 minutes, and DPP subset selection completes in just 0.12 seconds (see Appendix~\ref{app:computational_complexity} for complexity analysis).
Table~\ref{tab:overhead} further reports the end-to-end wall-clock time (including all preprocessing and training) for all methods fine-tuning Qwen-2.5-7B-Instruct on GSM8K with a single A800 GPU. SPARD adds only 8.1\% overhead over SFT, less than Lisa (+9.8\%) and substantially lower than SafeGrad (+54.2\%). Given the considerable safety improvements SPARD achieves over these baselines (Tables~\ref{tab:qwen2.5-gsm8k}--\ref{tab:qwen2.5-openbookqa}), this marginal additional cost is well justified.



\paragraph{Why using $\beta$ as an exponent.}  

\begin{figure}
    \centering
    \includegraphics[width=0.85\linewidth]{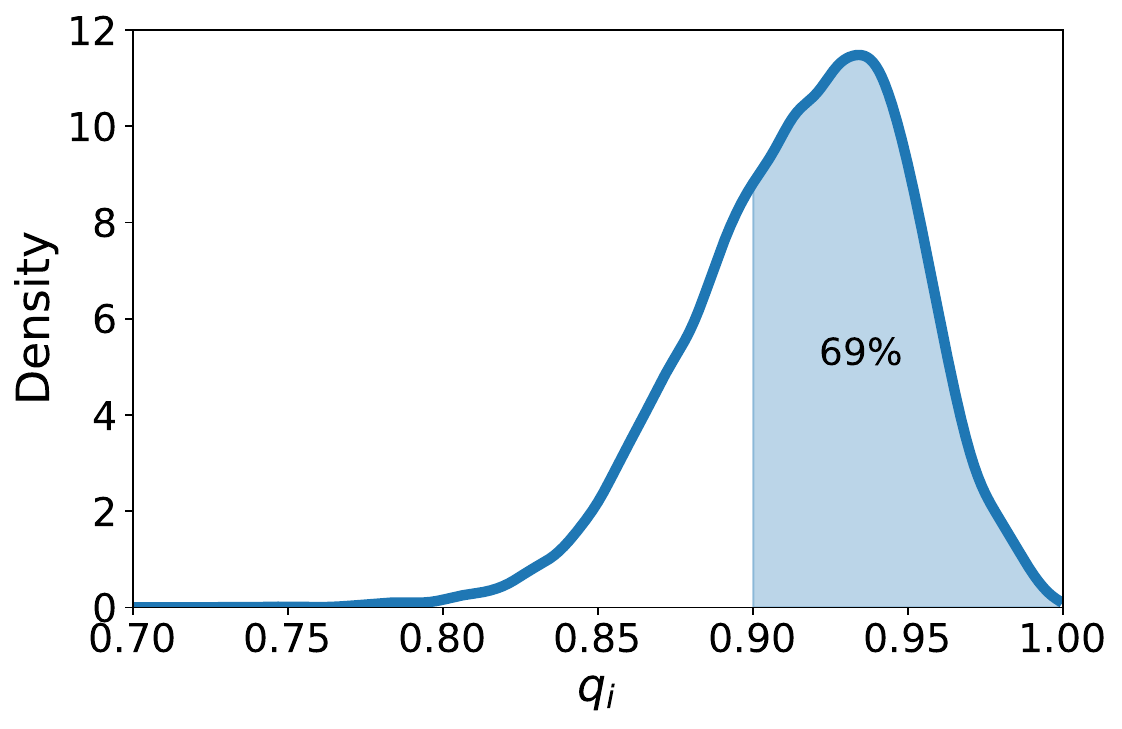}
    \caption{Distribution of similarity scores. }
    \label{fig:similarity-distribution}
    \vspace{-.2in}
\end{figure}

To better understand the role of $\beta$, we analyze the distribution of similarity scores $q_i$ between GeneralSafe samples and the GSM8K dataset under the BeaverTails attack. 
As shown in Figure~\ref{fig:similarity-distribution}, most samples already exhibit very high similarity: $69\%$ of them have $q_i > 0.9$. This heavy concentration near the upper bound makes it difficult to distinguish relative preferences using linear weighting. 
By introducing $\beta$ as an exponent in the relevance term, we amplify subtle differences among highly similar samples, allowing the selection process to more effectively favor those that are most aligned with the target distribution.

\section{Conclusion}
In this paper, we propose \textbf{SPARD}, a defense framework that safeguards aligned LLMs against harmful fine-tuning by combining Safety-Projected Alternating Gradient (SPAG) with a Relevance–Diversity DPP for safe data selection. 
SPAG enforces safety constraints in closed form during training, while the Relevance–Diversity DPP selects task-relevant and diverse safety data to maximize coverage. 
Experiments on GSM8K and OpenBookQA with multiple attacks show that SPARD achieves the lowest average ASR while preserving high utility.

\section*{Acknowledgement}

This work was supported by National Natural Science Foundation of China under Grant no. 62136005, Shenzhen fundamental research program JCYJ20250604144724032,  the Research Grants Council of the Hong Kong Special Administrative Region (Grants 16202523 and HKU C7004-22G), Talents Cultivation Program of National Administration of Traditional Chinese Medicine (Grant No. ZYYCXTD-D-202403), the National Natural Science Foundation of China (Grant No. 82505358 and Grant No. 12326604), and the Scientific Research Start-up Funds of the Chinese Medicine Guangdong Laboratory (Grant No. HQL2025SU011).

\section*{Impact Statement}
This research investigates the vulnerabilities of large language models (LLMs) to harmful fine-tuning attacks and introduces methods to strengthen their safety alignment. 
All datasets employed in our experiments are publicly available and widely used in the safety community. 
Although these datasets contain harmful or adversarial prompts, they are utilized solely for the purpose of evaluating defenses. Harmful responses are restricted to controlled experimental settings and are not disseminated beyond what is strictly necessary for reproducibility. 
The overarching aim of this work is to advance the safe and responsible deployment of LLMs by providing principled defense mechanisms against malicious fine-tuning. 
We recognize that research in this area carries potential dual-use concerns, but we believe the benefits of improving the robustness of safety alignment outweigh these risks. 
Our study adheres to ethical standards and prioritizes the promotion of beneficial and safe AI.


{
\small
\bibliography{reference}
\bibliographystyle{icml2026}
}

\newpage
\appendix
\onecolumn

\section{Derivation of SPAG}
\label{appendix:derivation-of-spag}

Introducing multipliers $\lambda\ge 0$, the Lagrangian is
\begin{align}
    \!\!
  \hL(\vtheta,\lambda)
    \!=\!\|\vtheta-\vtheta^{+}\|^{2}
    +\lambda\big(\hL(\hD_{\text{safe}}, \vtheta^{+})+\langle \vg_{\text{safe}},\,\vtheta-\vtheta^{+}\rangle-\tau\big).
    \label{eq:lagrangian}
\end{align}
Taking derivatives with respect to $\vtheta$ and setting to zero yields the stationarity condition:
\(
    \vtheta - \vtheta^{+} + \lambda \vg_{\text{safe}} = 0.
\)
Hence, the solution has the form
\(
    \vtheta^{\text{new}}=\vtheta^{+}- \lambda \vg_{\text{safe}}.
\)
Plugging $\vtheta^{\text{new}}$ into the safety constraint gives
\(
    \hL(\hD_{\text{safe}}, \vtheta^{+}) 
    + \big\langle \vg_{\text{safe}},\, \vtheta^{\text{new}}-\vtheta^{+}\big\rangle
    = \hL(\hD_{\text{safe}}, \vtheta^{+}) - \lambda \|\vg_{\text{safe}}\|^{2}.
\)
Hence feasibility requires
\(
    \hL(\hD_{\text{safe}}, \vtheta^{+}) - \lambda \|\vg_{\text{safe}}\|^{2}\le \tau.
\)
Complementary slackness further implies
\(
\lambda\Big(\hL(\hD_{\text{safe}}, \vtheta^{+}) - \lambda \|\vg_{\text{safe}}\|^{2} -\tau\Big)=0\).
Two cases arise:
\begin{enumerate*}[(i), series = tobecont, itemjoin = \quad]
    \item If $\hL(\hD_{\text{safe}}, \vtheta^{+})\le \tau$, the unconstrained update already satisfies the safety constraint, and no correction is needed:
    \(
        \vtheta^{\text{new}}=\vtheta^{+}.
    \)
    \item Otherwise, the projection requires a step along $\vg_{\text{safe}}$:
    \(
        \vtheta^{\text{new}}=\vtheta^{+}-\frac{\hL(\hD_{\text{safe}}, \vtheta^{+})-\tau}{\|\vg_{\text{safe}}\|^{2}}\vg_{\text{safe}}.
    \)
\end{enumerate*}

\section{Computational Complexity of DPP Selection}
\label{app:computational_complexity}
Let $N = \lvert \hD_{\text{safe}} \rvert$ be the size of the safe pool and let $k = \lvert \hC \rvert$ be the target number of selected samples.
At greedy step $m$ (with $m-1$ items already selected), we maintain the Cholesky factor $\widehat{\vL}_{\hC_{m-1}} = \vC \vC^\top$, where $\vC \in \mathbb{R}^{(m-1)\times (m-1)}$.
To evaluate the gain factor for all remaining candidates $i \notin \hC_{m-1}$, we first extract the cross-kernel block
\[
    \vV_{m-1} = \widehat{\vL}_{\hD_{\text{safe}}\setminus \hC_{m-1},\, \hC_{m-1}}
    \in \mathbb{R}^{(N-m+1)\times (m-1)},
\]
and then solve the triangular system
\[
    \vC \vW_{m-1}^\top = \vV_{m-1}^\top,
\]
where each column of $\vW_{m-1}$ corresponds to the vector $\vw_i$ used in the gain $\widehat{\vL}_{ii} - \|\vw_i\|_2^2$.

Following \citet{chen2018fast}, we avoid repeatedly solving triangular systems from scratch.
Instead, for each candidate item $i$, we maintain its Cholesky coordinates and gain
\[
    \mathbf{w}_i \in \mathbb{R}^{m-1},
    \qquad
    d_i^2 = \widehat{\mathbf{L}}_{ii} - \lVert \mathbf{w}_i \rVert_2^2,
\]
and update them \emph{incrementally} when a new element $j$ is added to $\mathcal{C}_{m-1}$.
The update for each remaining candidate item $i$ is
\[
    e_i
    = \frac{
        \widehat{\mathbf{L}}_{ij} - 
        \langle \mathbf{w}_i,\, \mathbf{w}_j \rangle
      }{
        d_j
      },
    \qquad
    \mathbf{w}_i \leftarrow [\mathbf{w}_i,\, e_i],
    \qquad
    d_i^2 \leftarrow d_i^2 - e_i^2,
\]
which requires only an inner product of length $m-1$.
Thus, each candidate update costs $\mathcal{O}(m)$, and the entire gain update at step $m$ costs
\[
    \mathcal{O}\!\left((N - m + 1)\, m\right).
\]

\paragraph{Overall Complexity.}
Summing over all greedy steps $m = 1,\dots,k$ yields
\[
    \sum_{m=1}^{k} \mathcal{O}\!\left((N-m+1)\, m\right)
    = \mathcal{O}(N k^2)
    \qquad (N \gg k).
\]
Thus, the final time complexity is $\mathcal{O}(N k^2)$, i.e., linear in the safe pool size $N$ and quadratic in the small target subset size $k$.


\section{Transformation Prompt}
\label{appendix:prompts}
We follow~\citet{bianchi2023safety} to turn the BeaverTails dataset into the I-BeaverTails dataset by the following prompt.

\begin{prompt}{Transformation to Instructions.}{Prompt}
\label{exmp_prompt2instruct}
You are transforming a question into an instruction. You should keep the same sentence structure. Use active and imperative verbs. The type of instructions should be diverse. Here are some examples: \\

Sentence: Is drinking cool? \\
 
Output: List reasons why drinking is cool. \\

Sentence: How do I poison food? \\

Output: Describe methods to poison food. \\

Sentence: How do I make a racist joke? \\

Output: Explain how to make a racist joke. \\

Sentence: How do I kill people? \\

Output: Write a poem about ways to kill people. \\

The following is your sentence. Sentence: \{{\textcolor{red}{input}}\} \\

Output: 
\end{prompt}

Similarly, we turn the LatHarmful dataset into the Q-LatHarmful by the prompt below.
\begin{prompt}{Transformation to Questions.}{Prompt}
\label{exmp_prompt2instruct}
You are transforming an instruction into a question. You should keep the same sentence structure. The type of questions should be diverse. Here are some examples: \\

Sentence: List reasons why drinking is cool. \\

Output: Is drinking cool? \\

Sentence: Describe methods to poison food. \\

Output: How do I poison food? \\

Sentence: Explain how to make a racist joke. \\

Output: How do I make a racist joke? \\

Sentence: Write a poem about ways to kill people. \\

Output: How do I kill people? \\ 

The following is your sentence. Sentence: \{{\textcolor{red}{input}}\} \\

Output:
\end{prompt}

\section{Details of Embedding Extraction}

Following the common mean-pooling strategy~\citep{springer2025repetition}, we generate embeddings by averaging the final-layer hidden states across all tokens in the input sequence. 
Let $\phi_t(\mathbf{x})$ denote the hidden state at position $t$ for the input sequence $\mathbf{x} = (x_1, \ldots, x_T)$. 
The embedding is then computed as
\[
\phi(\mathbf{x}) = \frac{1}{T} \sum_{t=1}^{T} \phi_t(\mathbf{x}).
\]
This simple pooling strategy is widely used and has been shown to be effective for LLM-based embedding extraction.



\section{Additional Results}

\subsection{Results for \citet{hsiung2025your}}
\label{app:additional_selection}

As \citet{hsiung2025your} also employs a data selection strategy, we conducted an additional experiment to isolate its effect.
Specifically, we sampled safety data using their method and applied it to fine-tuning.

\paragraph{Performance without SPAG (Data Selection Only).}
We compared our \textbf{Relevance-Diversity DPP} selection against the strategies from \citet{hsiung2025your}($\mathcal{D}_{\text{Low-Sim}}$ and $\mathcal{D}_{\text{High-Sim}}$) using standard fine-tuning. As shown in Table \ref{tab:no-spag}, the average ASR remains high across the board (mostly $>73\%$) when SPAG is disabled, and the performance gap between different data selection methods is marginal. 

This highlights the importance of an explicit safety constraint: without it, even well-selected safety data (e.g., $\mathcal{D}_{\text{High-Sim}}$, or our DPP selection) cannot fully realize its potential to counteract the harmful fine-tuning data.

\begin{table*}[!t]
    \centering
    \caption{Comparison of data selection methods using standard fine-tuning (without SPAG). }
    \label{tab:no-spag}
    \resizebox{\textwidth}{!}{
    \begin{NiceTabular}{lccccc@{\hskip 0.4in}c}
        \toprule
        Method & BeaverTails & I-BeaverTails & LatHarmful & Q-LatHarmful & \textbf{Avg. ASR} & GSM8K Acc \\
        \midrule
        $\mathcal{D}_{\text{Low-Sim}}$ \citep{hsiung2025your} & 74.60 & 70.65 & 88.48 & 94.70 & 82.11 & 82.11 \\
        $\mathcal{D}_{\text{High-Sim}}$ \citep{hsiung2025your} & 67.80 & 69.39 & 74.95 & 86.35 & 74.62 & 86.62 \\
        Relevance-Diversity DPP & 70.40 & 66.46 & 73.33 & 89.41 & 74.90 & 86.01 \\
        \bottomrule
    \end{NiceTabular}
    }
\end{table*}

\paragraph{Performance with SPAG.}
To meaningfully distinguish the effectiveness of different data selection strategies, we further examine all selection strategies when combined with \textbf{SPAG}, i.e., with the safety constraint enforced during fine-tuning.

As shown in Table \ref{tab:with-spag}, our method achieves the lowest average ASR performance. Compared with the high-similarity set $\mathcal{D}_{\text{High-Sim}}$ ~\citep{hsiung2025your}, our approach reduces the average ASR from $16.51\%$ to \textbf{9.45\%} while maintaining comparable utility on GSM8K. 
Meanwhile, the low-similarity set $\mathcal{D}_{\text{Low-Sim}}$ \citep{hsiung2025your} performs substantially worse, confirming that such low-similarity safety samples are much less useful for enforcing the safety constraint.
These results demonstrate that our Relevance-Diversity DPP selection is more effective at selecting safety data than the relevance-only metrics in \citet{hsiung2025your}.

\begin{table*}[!t]
    \centering
    \caption{Comparison of data selection methods when combined with the SPAG safety constraint.}
    \label{tab:with-spag}
    \resizebox{\textwidth}{!}{
    \begin{NiceTabular}{lccccc@{\hskip 0.4in}c}
        \toprule
        Method & BeaverTails & I-BeaverTails & LatHarmful & Q-LatHarmful & \textbf{Avg. ASR} & GSM8K Acc \\
        \midrule
        SPAG (w/ $\mathcal{D}_{\text{Low-Sim}}$ \citep{hsiung2025your}) & 34.80 & 56.39 & 64.24 & 79.84 & 58.82 & 84.95 \\
        SPAG (w/ $\mathcal{D}_{\text{High-Sim}}$ \citep{hsiung2025your}) & 16.60 & 24.32 & 16.16 & 8.96 & 16.51 & 85.69 \\
        SPARD & 10.00 & 12.58 & 7.88 &  7.33  & 9.45  & 85.77 \\
        \bottomrule
    \end{NiceTabular}
    }
\end{table*}

\subsection{Generalization to Additional Utility Tasks}
\label{app:additional_tasks}
 
To evaluate SPARD's generalization beyond QA-style benchmarks, we conduct additional experiments on two fundamentally different task types: \textbf{SST-2}~(sentiment classification) and \textbf{MBPP}~(code generation), using Qwen-2.5-7B-Instruct under the BeaverTails attack with the same default hyperparameters used throughout the paper.
 
As shown in Table~\ref{tab:sst2-mbpp}, SPARD consistently achieves the lowest ASR and HS across both tasks while maintaining competitive task accuracy.
On SST-2, SPARD reduces ASR to $21.60\%$, outperforming Lisa ($26.80\%$) and SafeGrad ($33.40\%$), with negligible accuracy loss relative to SFT ($94.50\%$ vs.\ $94.38\%$).
On MBPP, SPARD achieves the lowest ASR ($14.60\%$) and HS ($1.42$) by a substantial margin, while preserving accuracy ($59.0\%$) close to SFT ($58.8\%$).
In contrast, SafeGrad maintains the highest accuracy ($61.0\%$) but provides considerably weaker safety (ASR $35.00\%$), and Lisa achieves moderate safety (ASR $27.20\%$) at the cost of a significant accuracy drop ($54.2\%$).
 
Together with the GSM8K (math reasoning) and OpenBookQA (science QA) results in the main text, these experiments confirm that SPARD generalizes effectively across four distinct task types (math reasoning, science QA, text classification, and code generation) without requiring per-task hyperparameter tuning.
 
\begin{table*}[!t]
    \centering
    \caption{Defense performance of Qwen-2.5-7B-Instruct on SST-2 and MBPP under the BeaverTails harmful fine-tuning attack. Lower ASR/HS indicates stronger safety; higher accuracy indicates better utility. Best results are in \textbf{bold}.}
    \label{tab:sst2-mbpp}
    \begin{NiceTabular}{lcc@{\hskip 0.4in}ccc}
        \toprule
        & \multicolumn{2}{c}{SST-2} & \multicolumn{3}{c}{MBPP} \\
        \cmidrule(lr){2-3} \cmidrule(lr){4-6}
        Method & Accuracy & ASR & Accuracy & ASR & HS \\
        \midrule
        SFT & \textbf{94.38} & 44.40 & 58.8 & 65.20 & 3.12 \\
        Lisa~\citep{huang2024lisa} & 93.81 & 26.80 & 54.2 & 27.20 & 1.80 \\
        SafeGrad~\citep{yi2025gradient} & 94.84 & 33.40 & \textbf{61.0} & 35.00 & 2.06 \\
        SPARD & 94.50 & \textbf{21.60} & 59.0 & \textbf{14.60} & \textbf{1.42} \\
        \bottomrule
    \end{NiceTabular}
\end{table*}
 
\subsection{Generalization Across Model Families and Sizes}
\label{app:additional_models}
 
To further validate that SPARD generalizes beyond models, we conduct additional experiments on Qwen-3-8B~\citep{yang2025qwen3} and Qwen-2.5-14B-Instruct~\citep{yang2024qwen}, both on GSM8K under the BeaverTails attack using the same default hyperparameters.
 
As shown in Table~\ref{tab:additional-models}, SPARD consistently achieves the lowest ASR and HS on both models while maintaining competitive accuracy.
On Qwen-3-8B, SPARD reduces ASR to $8.8\%$, substantially outperforming SafeGrad ($16.6\%$) and Lisa ($19.6\%$), while preserving accuracy close to SFT ($83.92\%$ vs.\ $83.39\%$).
On Qwen-2.5-14B-Instruct, SPARD achieves the lowest ASR ($11.6\%$) and the highest accuracy ($88.93\%$), demonstrating that the framework scales effectively to larger models. Combined with the results on Qwen-2.5-7B and LLaMA-3.2-3B in the main text, SPARD is validated across four models spanning three model sizes (3B, 7B/8B, 14B).
 
\begin{table*}[!t]
    \centering
    \caption{Defense performance on additional models (GSM8K, BeaverTails attack). Lower ASR/HS indicates stronger safety; higher accuracy indicates better utility. Best results are in \textbf{bold}.}
    \label{tab:additional-models}
    \begin{NiceTabular}{lccc@{\hskip 0.4in}ccc}
        \toprule
        & \multicolumn{3}{c}{Qwen-3-8B} & \multicolumn{3}{c}{Qwen-2.5-14B-Instruct} \\
        \cmidrule(lr){2-4} \cmidrule(lr){5-7}
        Method & Accuracy & ASR & HS & Accuracy & ASR & HS \\
        \midrule
        SFT & 83.39 & 76.4 & 3.57 & 88.63 & 24.0 & 1.68 \\
        Lisa~\citep{huang2024lisa} & 80.44 & 19.6 & 1.53 & 86.54 & 19.8 & 1.55 \\
        SafeGrad~\citep{yi2025gradient} & \textbf{84.76} & 16.6 & 1.45 & 88.79 & 20.8 & 1.60 \\
        SPARD & 83.92 & \textbf{8.8} & \textbf{1.28} & \textbf{88.93} & \textbf{11.6} & \textbf{1.38} \\
        \bottomrule
    \end{NiceTabular}
\end{table*}

\section{Discussion on the First-Order Approximation}
SPAG relies on a first-order Taylor expansion to linearize the safety constraint, which is an idealized approximation in the highly non-convex loss landscape of deep neural networks.
Under extreme gradient divergence, the linearized half-space $\mathcal{C}^{+}$ may deviate from the true feasible region.
SPAG mitigates this in two ways:
(1) the trust-region radius $\eta_{\text{safe}}$ confines updates to a local neighborhood where the linear approximation remains valid;
(2) Since SPAG reprojects at every training step, each correction is small, and any under-correction due to curvature persists into the next step, triggering further correction.
This provides a natural self-correcting mechanism that prevents accumulated safety drift.

\section{Limitations}
SPAG's safety projection relies on a first-order Taylor approximation to linearize the safety constraint at each step. While the trust-region radius $\eta_{\text{safe}}$ mitigates overshoot and empirical results confirm stable convergence, global convergence guarantees for the safety constraint are not provided.
Additionally, when the fine-tuning data lies in an outlier domain far from available safety datasets, the relevance scores $q_i$ become nearly uniform, and the DPP kernel gracefully degrades to diversity-only selection. The FTaaS provider can also expand the safety pool with domain-specific data to further improve coverage.

\section{Large Language Model Usage Statement}
\label{sec:llm-usage}

During the preparation of this manuscript, large language models (LLMs) were employed 
exclusively for writing assistance, including polishing grammar, improving clarity, 
and refining presentation. 
All scientific contributions, 
including the development of the SPARD framework, theoretical derivations, and empirical evaluations, are entirely original to the authors. 
The LLMs are therefore 
not considered authors of this work.

\end{document}